\documentclass[sigconf,nonacm]{acmart}
\AtBeginDocument{%
  \providecommand\BibTeX{{%
    \normalfont B\kern-0.5em{\scshape i\kern-0.25em b}\kern-0.8em\TeX}}}

\usepackage{hhline}
\usepackage{multirow}
\usepackage{colortbl}
\usepackage{subfigure}
\usepackage{graphicx}
\usepackage{hyperref,xcolor}
\usepackage{algpseudocode}
\usepackage{amsmath}

\usepackage[ruled]{algorithm2e}

\AtBeginDocument{\hypersetup{pdfborderstyle={/S/S/W 1}}}

\begin{document}

\title{iPOF: An Extremely and Excitingly Simple Outlier Detection Booster via Infinite Propagation}

\author{Sibo Zhu}
\affiliation{%
  \institution{Brandeis University}
  \city{Waltham}
  \state{MA}
  \country{USA}
}
\email{siboz@brandeis.edu}

\author{Handong Zhao}
\affiliation{%
  \institution{Adobe Research}
  \city{San Jose}
  \state{CA}
  \country{USA}
}
\email{hazhao@adobe.com}

\author{Hongfu Liu}
\affiliation{%
  \institution{Brandeis University}
  \city{Waltham}
  \state{MA}
  \country{USA}
}
\email{hongfuliu@brandeis.edu}

\newcommand{\td}{{\bf\color{red} FIXME~}}
\newcommand{\handong}[1]{ {\color{blue} {[ HD:#1]}} }

\begin{abstract}
Outlier detection is one of the most popular and continuously rising topics in the data mining field due to its crucial academic value and extensive industrial applications. Among different settings, unsupervised outlier detection is the most challenging and practical one, which attracts tremendous efforts from diverse perspectives. In this paper, we consider the score-based outlier detection category and point out that the performance of current outlier detection algorithms might be further boosted by score propagation. Specifically, we propose Infinite Propagation of Outlier Factor (iPOF) algorithm, an extremely and excitingly simple outlier detection booster via infinite propagation. By employing score-based outlier detectors for initialization, iPOF updates each data point's outlier score by averaging the outlier factors of its nearest common neighbors. Extensive experimental results on numerous datasets in various domains demonstrate the effectiveness and efficiency of iPOF significantly over several classical and recent state-of-the-art methods. We also provide the parameter analysis on the number of neighbors, the unique parameter in iPOF, and different initial outlier detectors for general validation. It is worthy to note that iPOF brings in positive improvements ranging from 2\% to 46\% on the average level, and in some cases, iPOF boosts the performance over 3000\% over the original outlier detection algorithm. 
\end{abstract}


% \ccsdesc[300]{Computing methodologies~ Anomaly detection: \textit{Unsupervised outlier detection}}
% \ccsdesc{placeholder}


\maketitle

\section{INTRODUCTION}
Outlier detection, also known as anomaly detection, aims to identify the minority of data points with divergent characters from the majority. Due to its industrial value, there are numerous widely real-world applications of outlier detection, including credit card fraud, network intrusion, precision marketing, gene mutation, and so on. As an active research area, tremendous exploration has been taken to thrive outlier detection area with multiple practical settings to meet different real-world scenarios including supervised~\textcolor{gray}{\cite{aggarwal2013outlier,chalapathy2019deep}}, unsupervised~\textcolor{gray}{\cite{Breunig00SIR,gupta2014outlier,agrawal2015survey}}, even knowledge transferred settings~\textcolor{gray}{\cite{yu2019knowledge}}.

Among the above settings, unsupervised outlier detection is the most challenging one due to the nature of no external guidance information. Many algorithms have been proposed based on the various assumptions on the differences between inliers and outliers. Generally speaking, these algorithms can be roughly divided into crisp or soft categories. The methods in the crisp category explicitly provide the outlier membership with zero or one indicators. The representative method is K-means{-}{-}~\textcolor{gray}{\cite{chawla2013k}} with user pre-defined $o$ and $K$ denoting the numbers of outliers and clusters. It detects $o$ outliers and partitions the rest points into $K$ clusters, where the instances with large distances to the nearest centroid are regarded as outliers during the clustering process. COR~\textcolor{gray}{\cite{liu2019clustering}}, a variant of K-means{-}{-}, conducts joint clustering and outlier detection in the partition space. Differently, the soft category calculates a continuous score for each data point as the degree of outlierness, where the high value of a data point indicates the high probability of being an outlier. Then top-$K$ data points with the largest scores are regarded as outlier candidates. In this category, various methods are put forward from different assumptions or aspects, including density-based LOF~\textcolor{gray}{\cite{Breunig00SIR}}, COF~\textcolor{gray}{\cite{Tang02PKDD}}, distance-based LODF~\textcolor{gray}{\cite{Zhang09PKDD}}, angle-based FABOD~\textcolor{gray}{\cite{pham2012near}}, ensemble-based isolation Forest (iForest)~\textcolor{gray}{\cite{LiuFei2008}}, eigenvector-based OPCA~\textcolor{gray}{\cite{lee2012anomaly}}, cluster-based TONMF~\textcolor{gray}{\cite{Kannan17SDM}}, deep learning-based anomalous
event detection~~\textcolor{gray}{\cite{xu2015learning}}, and so on. More details on outlier detection can be found in the surveys~\textcolor{gray}{\cite{gupta2014outlier,agrawal2015survey}}. 

In this paper, we consider the score-based unsupervised outlier detection category, where the Local Outlier Factor (LOF) is one of the most popular approaches. LOF measures the local deviation of a target data point with respect to its neighbors. In other words, the outlier score of the target data point in LOF depends on the local density of $K$ nearest neighbors, rather than the density of the target data point. From this view, we regard LOF as a one-round outlier factor propagation based on local neighborhood structure. It is interesting to see whether multi-round propagation will increase the gap between inliers and outliers, the final score will converge with infinite propagation, and the propagation will benefit other outlier detectors. In light of this, we propose infinite Propagation of Outlier Factor (iPOF) to address the above questions. iPOF is based on local neighborhood structure and assumes that inliers' friends are more likely to be inliers, and inliers refuse to make friends with outliers. Here the friendship is defined based on the common neighbour. In Figure\textcolor{red}{~\ref{fig:neighbor}}, $B$ and $D$ are the nearest neighbors of $F$; however, $B$ and $D$'s nearest neighbors do not include $F$, which means that $B$ and $D$ do not want to make friends with $F$. In light of this, we build a common neighborhood graph based on the local structure for the outlierness propagation. iPOF utilizes the outlier scores initialized by some outlier detector, such as LOF or other outlier detectors, and iteratively updates the outlier scores by averaging the scores of local neighbors. As a post-processing technique, iPOF can further boost the existing score-based outlier detectors via averaging propagation until convergence. Honestly, iPOF is extremely and excitingly simple, but quite effective. Extensive experiments demonstrate that iPOF enhances several outlier detectors by increasing the improvements from 2\% to 46\% on average level. The major contributions of iPOF are summarized as follows:

\begin{figure}[t]
  \centering
    \includegraphics[width=0.4\textwidth]{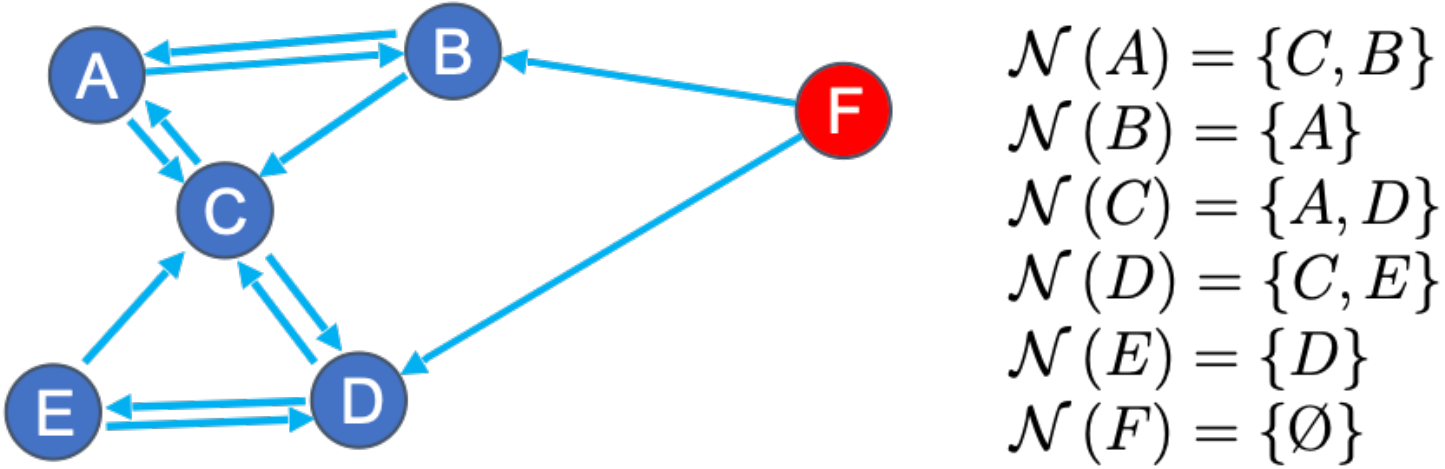}
  \caption{A toy example of common neighbors, where the blue nodes denote the inliers, and the red node represents the outlier. The directed edges denote the neighborhood relationship. For instance, the node $B$ and $D$ are 2 nearest neighbors of $F$, which indicates the edges are directed from $F$ to $B$ and $D$. The common neighbor sets of these seven nodes are given on the right side. } \label{fig:neighbor}\vspace{-4mm}
\end{figure}

\begin{itemize}
    \item We propose infinite Propagation of Outlier Factor (iPOF), a performance-boosting technique for existing score-based outlier detection algorithms. iPOF employs the local neighbor structure to propagate the outlier scores to converge the scores of outliers and inliers for better distinguishing. 
    \item Technically, iPOF only requests nearest neighbor calculation and averaging, where the nearest neighbor information can be re-used from the density- or distance-based initial outlier detectors. Hence, iPOF is extremely and excitingly simple, time- \& space-efficient and easy to implement in parallel. 
    \item Experiments on 17 real-world datasets in diverse domains demonstrate the effectiveness of iPOF over numerous classical and recent outlier detectors. In some cases, iPOF brings in over 3000\% improvements over the original detector.
\end{itemize}

The rest of this paper is organized as follows. Section\textcolor{red}{~\ref{sec:related}} introduces the related work of outlier detection in terms of supervised, semi-supervised and unsupervised setting. Section\textcolor{red}{~\ref{sec:motivation}} provides an illustrative example to deliver the idea of iPOF, and the core principle and algorithm of iPOF is elaborated in Section\textcolor{red}{~\ref{sec:method}}. We conduct comprehensive experiments on algorithmic comparison with other outlier detection algorithms and in-depth factor exploration in Section\textcolor{red}{~\ref{sec:experiment}}. Finally, Section\textcolor{red}{~\ref{sec:conclusion}} draws conclusion to the whole paper.

\section{RELATED WORK}\label{sec:related}
In this section, we introduce the related work of outlier detection according to the label availability and highlight the differences between existing works and our proposed iPOF.

\textit{Supervised Outlier Detection}. With enough labeled data, supervised outlier detection is a binary classification problem in essence, where a bunch of labeled inlier and outlier samples is employed to build a predictive model for new sample prediction~\textcolor{gray}{\cite{aggarwal2013outlier,gornitz2013toward,gogoi2010anomaly}}. Different from the conventional classification problem, the challenge of supervised outlier detection lies in the imbalance of inlier and outlier class sizes. Many balancing techniques including sampling~\textcolor{gray}{\cite{yang2013classification}}, re-weighting~\textcolor{gray}{\cite{ren2018learning}}, deep representation learning~\textcolor{gray}{\cite{chalapathy2019deep,kwon2017survey,li2017transferred}} can be used to tackle the supervised outlier detection.

\textit{Semi-supervised Outlier Detection}. Semi-supervised learning uses both unlabeled and labeled data for particular tasks. It is worthy to note that only inlier samples are available in the training data in the setting of semi-supervised outlier detection. The most representative algorithm is one-class SVM~\textcolor{gray}{\cite{ma2003time,erfani2016high}}, which seeks several support vectors to describe the boundary of the inlier class for the unseen new sample prediction. This idea is extended to the kernel version~\textcolor{gray}{\cite{roth2005outlier,gautam2019localized}}, and deep representation~\textcolor{gray}{\cite{chalapathy2018anomaly,sabokrou2018adversarially,wulsin2010semi,akcay2018ganomaly}} as well.

\begin{figure*}[htb]
  \centering
    \subfigure[\textit{Data Points Visualization}]{
    \includegraphics[width=0.34\textwidth]{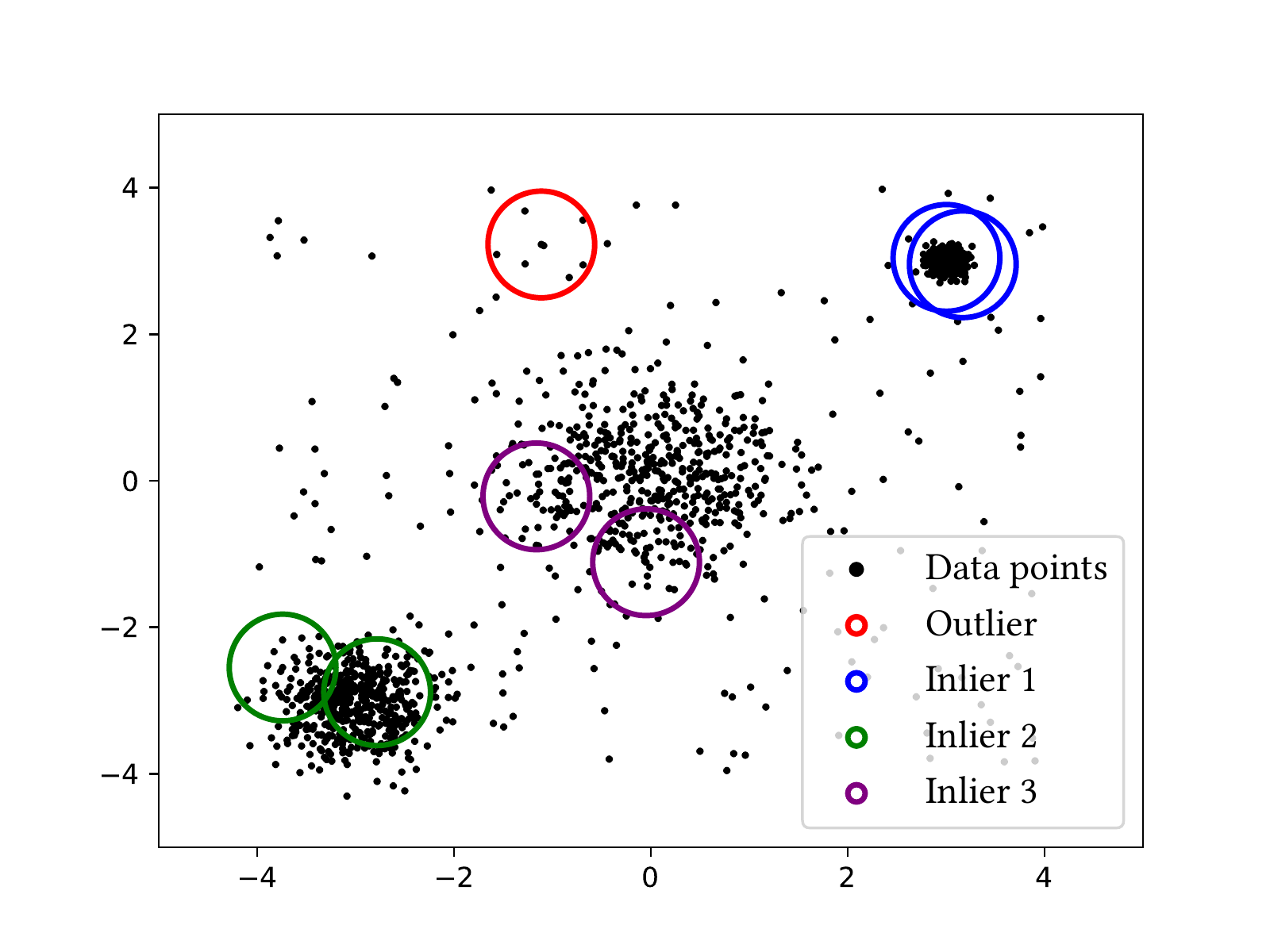}\label{fig:data-dis}}\hspace{-6mm}
    \subfigure[\textit{Scores of outliers and inliers via iPOF}]{
    \includegraphics[width=0.34\textwidth]{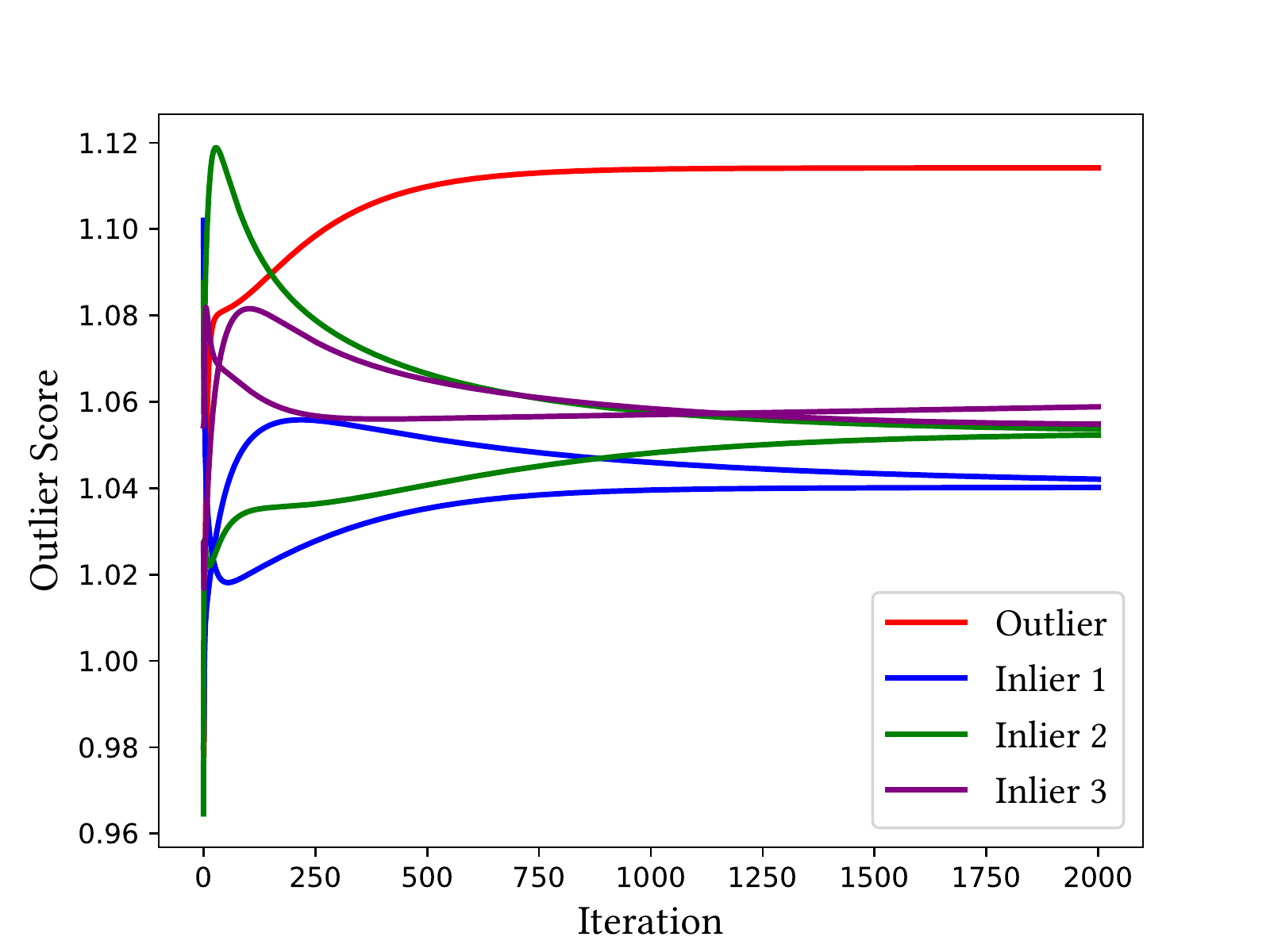}\label{fig:data-obs}}\hspace{-6mm}
    \subfigure[\textit{Performance of iPOF and LOF during iterations}]{
    \includegraphics[width=0.34\textwidth]{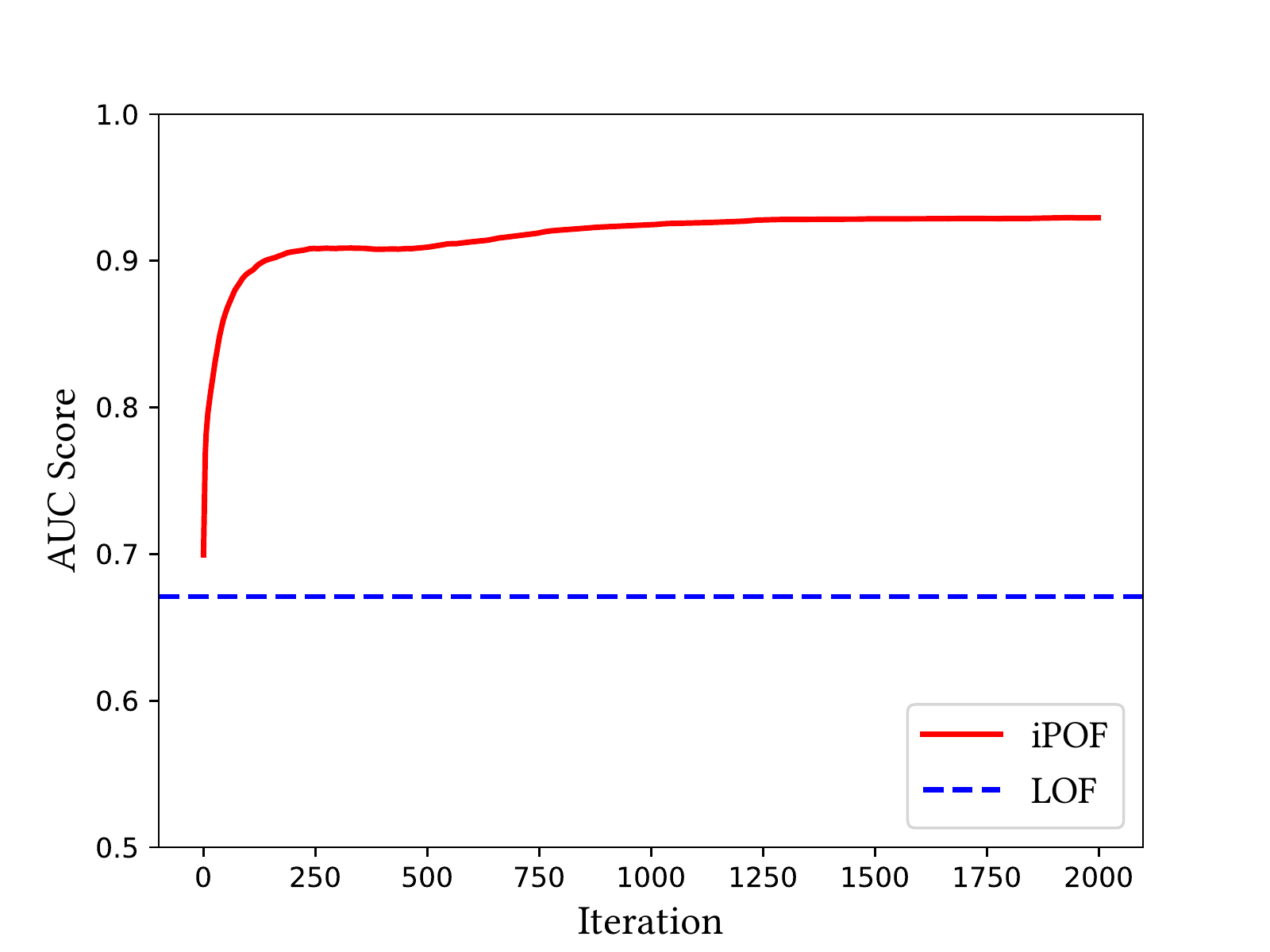}\label{fig:auc-record}}
  \caption{iPOF on a 2D synthetic dataset. (a) shows the data distribution with three different density Gaussian clusters of 500 data points each and 150 outliers samples that are randomly landed out of these clusters; (b) shows the outlier score trend with respect to the iteration of iPOF on several inliers and one outlier that highlighted by different colorful circles in (a); and (c) demonstrates the AUC scores of iPOF during iterations, where iPOF initializes by LOF and increases the AUC score from 0.6709 to 0.9294. The improvement of iPOF over LOF on this synthetic dataset is up to 38\%.} \label{fig:motivation}
\end{figure*}

\textit{Unsupervised Outlier Detection}. Among the diverse settings of outlier detection, the unsupervised scenario is the most challenging and practical one, that cannot be solved by conventional classification techniques. Tremendous and sustaining efforts have been devoted to making unsupervised outlier detection rich and contending. Generally speaking, the existing algorithms in this category can be divided into two groups by the type of outlier scores. K-means{-}{-}~\textcolor{gray}{\cite{chawla2013k}} and COR~\textcolor{gray}{\cite{liu2019clustering}} provide the binary scores for samples to identify outlier candidates, while most other detectors calculate a continuous score to denote the degree of outlierness. Based on different assumptions, numerous score-based outlier detectors have been proposed including density-based DBSCAN~\textcolor{gray}{\cite{ester1996density}}, Local Outlier Factor (LOF)~\textcolor{gray}{\cite{Breunig00SIR}}, Connectivity-based Outlier Factor (COF)~\textcolor{gray}{\cite{Tang02PKDD}}, distance-based Local Distance-based Outlier Factor (LODF)~\textcolor{gray}{\cite{Zhang09PKDD}}, frequent pattern-based Fp-outlier~\textcolor{gray}{\cite{he2005fp}}, angle-based ABOD~\textcolor{gray}{\cite{kriegel2008angle,pham2012near}}. Moreover, some studies purse outlier detection by representation learning as well from the subspace~\textcolor{gray}{\cite{10.1007/978-3-642-01307-2_86}}, low-rank~\textcolor{gray}{\cite{zhao2015dual}} and matrix-completion~\textcolor{gray}{\cite{Kannan17SDM}}, respectively. Since the above basic outlier detectors are highly assumption-dependent, some ensemble outlier detector algorithms are put forward to alleviate the negative impact of assumption including ensemble-based iForest~\textcolor{gray}{\cite{LiuFei2008}}, Bi-Sampling Outlier Detection (BSOD)~\textcolor{gray}{\cite{liu2016outlier}}, Feature Bagging~\textcolor{gray}{\cite{lazarevic2005feature}}, Locally Selective Combination of Parallel Outlier Ensembles (LSCP)~\textcolor{gray}{\cite{DBLP:journals/corr/abs-1812-01528}}, Combination of outlier detectors by taking the median of the scores (Median)~\textcolor{gray}{\cite{Aggarwal2015TheoreticalFA}}, Combination of outlier detectors by taking the Average of Maximum (AOM)~\textcolor{gray}{\cite{Aggarwal2015TheoreticalFA}} and Maximum of Average (MOA)~\textcolor{gray}{\cite{Aggarwal2015TheoreticalFA}}. Due to the superior performance of supervised learning over unsupervised learning, some studies transform the unsupervised outlier detection problem into a classification problem by artificially generating potential outliers. Typically, these algorithms first generate a set of potential outliers and treat them as labels. Then a classifier is trained for subsequent detection. Among them, the one-class classification method~\textcolor{gray}{\cite{hempstalk2008one}} generates informative potential outliers based on the probability density function of the real data. The distribution-based artificial anomaly method~\textcolor{gray}{\cite{fan2004using}} exploits a heuristic to randomly change the value of one feature
of an object by assuming that the boundary may be very close to the existing data. One-Class Random Forests~\textcolor{gray}{\cite{desir2013one}} makes use of classifier ensemble randomization principles for outlier generation procedure. Recently there is increasing attention focusing on generating fake outliers by min-max training with generative adversarial networks~\textcolor{gray}{\cite{liu2019generative,zenati2018efficient,li2018anomaly}}.

Different from the existing literature in the outlier detector area, our iPOF approach is a post-processing technique for boosting score-based outlier detector methods, rather than an outlier detector. iPOF is based on local neighborhood structure and assumes that an object is more likely to be an outlier if its neighbors are outliers as well, while an object is an inlier if it is surrounded by inliers. By literately averaging the outlier scores with neighbors, iPOF converges the scores of outliers and inliers for better distinguishing, hence increases the performance of unsupervised outlier detection.

\section{MOTIVATION}\label{sec:motivation}
Unsupervised outlier detection employs some criterion to seek outlier candidates deviation from major normal points without external supervision. Usually, each data point gets a score calculated to identify the degree of outlierness, where top $K$ points with the largest scores are regarded as outlier candidates. Tremendous efforts have been made in the outlier detection area based on different assumptions, including density, distribution, clustering, angle, and so on. Among these studies, Local Outlier Factor (LOF) is one of the most popular density-based methods, where outliers are identified by comparing the local density of the data point and its neighbors~\textcolor{gray}{\cite{Breunig00SIR}}. Then many variants including  Connectivity-based Outlier Factor (COF)~\textcolor{gray}{\cite{Tang02PKDD}}, Local Distance-based Outlier Factor (LODF)~\textcolor{gray}{\cite{Zhang09PKDD}} follow to further improve the performance of LOF. 

Here we take LOF as an example for further analysis that measures the local deviation of a target data point with respect to its neighbors. The outlier score of the target data point in LOF depends on the local density of $K$ nearest neighbors, rather than the density of the target data point. From this view, we regard LOF as a one-round outlier factor propagation based on local neighborhood structure. It is interesting to see whether multi-round propagation will increase the gap between inliers and outliers, and the final score will converge with infinite propagation. In this paper, we are trying to answer the questions of whether the performance of these outlier detectors can be further improved via a simple average propagation strategy based on local neighborhood structure.

To validate the above conjecture, we conduct the experiments on a synthetic dataset. Figure\textcolor{red}{~\ref{fig:motivation}} shows the data distribution, outlier scores of several selected data points, and the performance during the propagation of iPOF. The 2D dataset contains three Gaussian clusters of 500 objects, each with different densities and 150 outliers samples that are randomly landed out of these clusters. From Figure\textcolor{red}{~\ref{fig:data-obs}}, the original outlier scores by LOF at the $0$-th iteration mix the inlier and outlier samples. That is the intrinsic reason that LOF cannot well tackle this synthetic dataset. Instead, our iPOF method initializes with LOF value for each data point and propagates the scores through local neighborhood structure, which increases not only the gap between inliers and outliers but also narrows the score divergence of inliers from the same group. More technical details and properties will be provided in Section\textcolor{red}{~\ref{sec:method}}. There are two clear observations to validate the above point in Figure\textcolor{red}{~\ref{fig:data-obs}}. (1) The gap among the red curve and other curves are enlarged along with the iteration. (2) The curves with the same color denoting the inliers from the same cluster converge together even they have different initial outlier scores. For instance, two green curves present two inliers in the same cluster, where one is the high-density central area, another is located in the boundary of the cluster with relatively low density. Through the propagation of iPOF, their final scores achieve the same. Finally, Figure\textcolor{red}{~\ref{fig:auc-record}} demonstrates the AUC performance of iPOF. It is appealing to see that initialized by LOF, iPOF increases the AUC score from 0.6709 to 0.9294, bringing in over 38\% improvements on this synthetic dataset.

The illustrative example demonstrated in Figure\textcolor{red}{~\ref{fig:motivation}} shows that LOF can further gain significant improvements via simple local neighborhood average propagation. It is worthy to note that the local neighborhood structure is pre-calculated through the LOF, which indicates that our iPOF method boosts the performance only via iterative averaging. In the next section, we will provide more technical details and properties of our proposed iPOF method.

\begin{table}[!t]
\small
\caption{Notations and descriptions}
\vspace{-4mm}
\centering
\begin{tabular}{c| l} \hline
Notation &   Description \\ \hline 
$n$ & Number of instances \\
$k$ & Number of neighbors \\
$K$ & Number of neighbors for propagation\\
$x_l$ & $l$-th data point\\
$s_l$ & Outlier score of $x_l$\\
$\mathcal{N}_K(x_l)$ &  Top-$K$ points in the common neighbor set of $x_l$\\
\hline
\end{tabular}\vspace{-4mm}
\label{tab:notation}
\end{table}

\section{METHODOLOGY}\label{sec:method}
Generally speaking, outlier detectors predict outlier scores under various assumptions in terms of density, distribution, clustering, angle, and so on. Our proposed infinite Propagation of Outlier Factor (iPOF) based on local neighborhood structure, assuming that \textit{inliers' friends are more likely to be inliers, and inliers refuse to make friends with outliers; thus, outliers' friends are more likely to be outliers}. Here the friendship is defined based on the common nearest neighbour, which means that the direct neighborhood relationship is not friendship. 

Before illustrating the algorithm of iPOF, we first introduce the concept of common neighbor set with an illustrative example. Figure\textcolor{red}{~\ref{fig:neighbor}} shows a toy example of common neighbor with six blue inlier nodes and one red outlier node, where $k=2$ denotes that we calculate 2 nearest neighbors for every node. The directed edges denote the common neighborhood relationship. For instance, the node $B$ and $D$ are 2 nearest neighbors of $F$, which indicates the edges are directed from $F$ to $B$ and $D$. The \textit{common neighbor set} $\mathcal{N}$ of a target data point includes all the nodes pointing to the target one. Here $k$ is used to construct the network and does not mean that there are at most $k$ elements in the set. Given the fixed network derived by the local neighbors, iPOF iteratively updates the outlier score of each data point via simply averaging its top $K$ points scores in its common neighbor sets. In the case of Figure\textcolor{red}{~\ref{fig:neighbor}}, we first initialize the outlier score of each node by some outlier detectors, which leads that node $F$ has a relatively high score, and others have relatively low scores. Then we apply iPOF to propagate the outlier scores. Note that the in-degree of node $F$ is zero, which means its score will not get affected by other inlier nodes during the propagation. Instead, there exist several paths among these blue inlier nodes. With proper propagation, the outlier scores of these inliers will converge to the same value. Note that $k$ denotes the number of neighbors, while $K$ is the number of common neighbors for the propagation in iPOF. Note that $k<K$ and $k$ can be relatively large to incorporate more common neighbors, while $K$ is the key parameter to control the propagation. Table~\ref{tab:notation} shows the notations and descriptions used in this paper.

% To distinguish the scores of outliers and inliers, we build an common neighborhood graph based on the local structure for the outlierness propagation. To achieve this, iPOF utilizes the outlier scores initialized by some outlier detector, such as LOF, COF, and iteratively updates the outlier scores by averaging the scores of local neighbors. The above process is repeated until convergence. 

Given a dataset containing $n$ data points $x_l$, $1\leq l \leq n$, we first initialize the outlier score $s_l^{(0)}$ of each data point via some existing outlier detector. Then, we calculate the $k$-nearest neighbors based on the original data features for each data point and obtain the common neighbor sets, based on which our proposed iPOF propagates the outlier scores with top-$K$ points in the common neighbor sets as follows:
\begin{equation}\label{eq:update}
    s_l^{(t+1)} = (\sum_{j\in\mathcal{N}_K(x_l)} s_j^{(t)} + s_l^{(t)})/(|\mathcal{N}_K(x_l)|+1),
\end{equation}
where $\mathcal{N}_K(x_l)$ includes top-$K$ points in the common neighbor set of $x_l$, $t$ is the index of iteration. The outlier score of each point is iteratively updated by averaging the scores of common neighbors until convergence of all data points' scores. Therefore, the core idea of iPOF is the local average to converge outlier and inlier scores so that, in general, the outliers have high scores and inliers have low scores during the propagation. By this means, the gap between inliers and outliers can be further improved to distinguish them. The network constructed by local common neighbors naturally determines the outlier score flow among these data points and makes the outlier scores along a path equal when iPOF achieves convergence. By choosing top-$K$ common neighbors, iPOF splits up the network into several connected communities and the outliers. In general, iPOF is extremely and excitingly simple. We summarize the whole process of iPOF in Algorithm.\textcolor{red}{~\ref{alg:optimize}}. For convergence, it is easy to understand that iPOF is guaranteed to converge with any initialization, when the data points in a same connected community have the same score. 

\begin{algorithm}[t]
\caption{Infinite Propagation of Outlier Factor (iPOF)}\label{alg:optimize}
    \KwIn{Data points $x_l$, $1\leq l \leq n$;} 
    \ \ \ \ \ \ \ \ \ \ \  \ \ Number of common neighbors for propagation $K$.\\
    \KwOut{Outlier scores: $s_l^{(t)}$, $1\leq l \leq n$.} 
    1. Initialize the outlier score $s_l^{(0)}$ for each data point with some outlier detector;\\
    2. Build $K$-nearest common neighbor graph;\\
    3. Set $t=0$;\\
    4. \While{not converge}{
        $t=t+1$;\\
        For $x_l$, update the outlier score $s_l^{(t)}$ by Eq.~\textcolor{red}{\eqref{eq:update}};
    }
\end{algorithm}

\section{EXPERIMENTAL RESULTS}\label{sec:experiment}
In this section, we provide comprehensive experiments of iPOF, including algorithmic performance and in-depth exploration. We compare iPOF with 12 classical and recent unsupervised outlier detectors on 17 datasets in terms of quantitative performance and execution time. We also provide the parameter analysis on the number of neighbors and different initial outlier detectors in iPOF.

\begin{table}[t]
\caption{Characteristics of outlier detection datasets}\label{tab:datasets}
\vspace{-4mm}
\resizebox{.48\textwidth}{!}{
 \begin{tabular}{c c c c c c}
 \hline
 \textbf{Dataset} & \textbf{\#Size}  & \textbf{\#Dim} & \textbf{\#Outlier} & \textbf{Outliers\%} & \textbf{Outlier object}\\ [0.5ex] 
 \hline\hline
 \textit{Arrhythmia} & 452  & 274& 66 & 14.6\% & Affected patients\\ 
 \textit{BreastW} & 683  & 9& 239 & 35.0\% & Malignant cases\\
 \textit{Cardio} & 1831  & 21& 176 & 9.6\% & Patients\\
 \textit{Glass} & 214  & 9& 9 & 4.2\%& Glass identification\\
 \textit{Ionosphere} & 351  & 33& 126 &35.9\%& Radar data\\
 \textit{Mammography} & 11183  & 6& 260 &2.3\%& Calcification classes\\
 \textit{Mnist} & 7603  & 100& 700 &9.2\%& Handwritten digits\\
 \textit{Optdigits} & 5126  & 64& 150 &2.9\%& Handwritten digits\\
 \textit{Pendigits} & 6870  & 16& 156 &2.2\%& Handwritten digits\\
 \textit{Pima} & 768  & 8& 268&34.9\% & Patients\\
 \textit{Satellite} & 6435  & 36& 2036 &31.6\%& Satellite images\\
 \textit{Satimage-2} & 214  & 9& 9 &4.2\%& Satellite images\\
 \textit{Shuttle*} & 4909  & 9& 351& 7.1\%& Statlog points\\
 \textit{Speech} & 3686  & 400& 61 &1.7\%& Accents\\
 \textit{Vertebral} & 240  & 6& 30 &12.5\%& Patients\\
 \textit{Vowels} & 1456  & 12& 50&3.4\% & Speakers\\
 \textit{Wine} & 129  & 13& 10 &7.8\%& Wine\\ 
 \hline
 \multicolumn{6}{c}{Note: * means that we randomly sample 10\% data points from the whole set.}
\end{tabular}}\vspace{-4mm}
\end{table}

\begin{table*}[h!]
\caption {AUC results of iPOF and other detectors on 17 datasets} 

    \centering
    \resizebox{.98\textwidth}{!}{
    \begin{tabular}{|p{5cm}||p{5cm}|p{5cm}|p{5cm}|p{5cm}|p{5cm}|p{5cm}|p{5cm}|p{5cm}|p{5cm}|p{5cm}|p{5cm}|p{5cm}|p{5cm}}
    \cline{1-14}
    % \hline
    % \multicolumn{1}{c||}{} & \multicolumn{2}{c|}{\textbf{Dataset}} \\
    \cline{2-14}
    
    \multicolumn{1}{c||}{\textbf{Datasets (outliers\%)}} & 
    \multicolumn{1}{c|}{\textbf{iPOF}}& 
    \multicolumn{1}{c}{\textbf{LOF}}& 
    \multicolumn{1}{c}{\textbf{COF}}& 
    \multicolumn{1}{c}{\textbf{FABOD}}& 
    \multicolumn{1}{c|}{\textbf{SOD}}& 
    \multicolumn{1}{c}{\textbf{MGAAL}}&
    \multicolumn{1}{c||}{\textbf{SGAAL}}&
    \multicolumn{1}{c}{\textbf{iForest}}& 
    \multicolumn{1}{c}{\textbf{LSCP}}& 
    \multicolumn{1}{c}{\textbf{FB}}&  
    \multicolumn{1}{c}{\textbf{Median}}& 
    \multicolumn{1}{c}{\textbf{AOM}}& 
    \multicolumn{1}{c}{\textbf{MOA}}\\
    
    \hhline{=||=|====|==||======}
    \multicolumn{1}{c||}{Arrhythmia (15\%)} & \multicolumn{1}{c|}{0.6284} & \multicolumn{1}{c}{0.7319} & \multicolumn{1}{c}{0.7094} & \multicolumn{1}{c}{0.2315} & \multicolumn{1}{c|}{0.7201} & \multicolumn{1}{c}{0.5000} & \multicolumn{1}{c||}{0.5000} & \multicolumn{1}{c}{0.3267} & \multicolumn{1}{c}{0.5975} & \multicolumn{1}{c}{\textbf{0.7953}} & \multicolumn{1}{c}{0.7744} & \multicolumn{1}{c}{0.7699} & \multicolumn{1}{c}{0.7714}\\
    
    \multicolumn{1}{c||}{BreastW (35\%)} & \multicolumn{1}{c|}{0.5481} & \multicolumn{1}{c}{0.4106} & \multicolumn{1}{c}{0.4466} &   \multicolumn{1}{c}{N/A} & \multicolumn{1}{c|}{0.9374} & \multicolumn{1}{c}{\textbf{0.9883}} & \multicolumn{1}{c||}{0.9882} & \multicolumn{1}{c}{0.0150} & \multicolumn{1}{c}{0.5479} & \multicolumn{1}{c}{0.2358} & \multicolumn{1}{c}{0.6606} & \multicolumn{1}{c}{0.5095} & \multicolumn{1}{c}{0.4943}\\
    
    \multicolumn{1}{c||}{Cardio (9.6\%)} & \multicolumn{1}{c|}{0.6564} & \multicolumn{1}{c}{0.5968} & \multicolumn{1}{c}{0.5641} &   \multicolumn{1}{c}{0.4126} & \multicolumn{1}{c|}{0.5617} & \multicolumn{1}{c}{0.3778} & \multicolumn{1}{c||}{0.3746} & \multicolumn{1}{c}{0.2396} & \multicolumn{1}{c}{0.5441} & \multicolumn{1}{c}{0.5966} & \multicolumn{1}{c}{0.8660} & \multicolumn{1}{c}{\textbf{0.9029}} & \multicolumn{1}{c}{0.8905}\\
    
    \multicolumn{1}{c||}{Glass (4.2\%)} & \multicolumn{1}{c|}{\textbf{0.9222}} & \multicolumn{1}{c}{0.7827} & \multicolumn{1}{c}{0.7572} &   \multicolumn{1}{c}{0.1463} & \multicolumn{1}{c|}{0.7702} & \multicolumn{1}{c}{0.7035} & \multicolumn{1}{c||}{0.7038} & \multicolumn{1}{c}{0.3707} & \multicolumn{1}{c}{0.5268} & \multicolumn{1}{c}{0.8244} & \multicolumn{1}{c}{0.7089} & \multicolumn{1}{c}{0.7415} & \multicolumn{1}{c}{0.7881}\\
    
    \multicolumn{1}{c||}{Ionosphere (36\%)} & \multicolumn{1}{c|}{0.8859} & \multicolumn{1}{c}{0.9023} & \multicolumn{1}{c}{0.9103} &   \multicolumn{1}{c}{0.0720} & \multicolumn{1}{c|}{0.8948} & \multicolumn{1}{c}{0.4270} & \multicolumn{1}{c||}{0.4273} & \multicolumn{1}{c}{0.1549} & \multicolumn{1}{c}{0.6327} & \multicolumn{1}{c}{0.8758} & \multicolumn{1}{c}{\textbf{0.9193}} & \multicolumn{1}{c}{0.9038} & \multicolumn{1}{c}{0.8972}\\
    
    \multicolumn{1}{c||}{Mammography (2.32\%)} & \multicolumn{1}{c|}{\textbf{0.8575}} & \multicolumn{1}{c}{0.6709} & \multicolumn{1}{c}{0.6863} &   \multicolumn{1}{c}{N/A} & \multicolumn{1}{c|}{0.7977} & \multicolumn{1}{c}{0.7380} & \multicolumn{1}{c||}{0.7429} & \multicolumn{1}{c}{0.3072} & \multicolumn{1}{c}{N/A} & \multicolumn{1}{c}{0.7189} & \multicolumn{1}{c}{0.8099} & \multicolumn{1}{c}{0.8279} & \multicolumn{1}{c}{0.8259}\\
    
    \multicolumn{1}{c||}{Mnist(9.2\%)} & \multicolumn{1}{c|}{\textbf{0.9232}} & \multicolumn{1}{c}{0.5824} & \multicolumn{1}{c}{0.5543} &   \multicolumn{1}{c}{0.2435} & \multicolumn{1}{c|}{0.5902} & \multicolumn{1}{c}{0.4993} & \multicolumn{1}{c||}{0.5000} & \multicolumn{1}{c}{0.2224} & \multicolumn{1}{c}{0.5683} & \multicolumn{1}{c}{0.6518} & \multicolumn{1}{c}{0.8220} & \multicolumn{1}{c}{0.8484} & \multicolumn{1}{c}{0.8379}\\
    
    \multicolumn{1}{c||}{Optdigits (3\%)} & \multicolumn{1}{c|}{\textbf{0.9676}} & \multicolumn{1}{c}{0.6154} & \multicolumn{1}{c}{0.5720} &   \multicolumn{1}{c}{0.4893} & \multicolumn{1}{c|}{0.5637} & \multicolumn{1}{c}{0.5512} & \multicolumn{1}{c||}{0.5833} & \multicolumn{1}{c}{0.2776} & \multicolumn{1}{c}{0.6088} & \multicolumn{1}{c}{0.5722} & \multicolumn{1}{c}{0.4521} & \multicolumn{1}{c}{0.5865} & \multicolumn{1}{c}{0.5347}\\
    
    \multicolumn{1}{c||}{Pendigits (2.27\%)} & \multicolumn{1}{c|}{\textbf{0.9562}} & \multicolumn{1}{c}{0.5256} & \multicolumn{1}{c}{0.5314} &   \multicolumn{1}{c}{0.3208} & \multicolumn{1}{c|}{0.6753} & \multicolumn{1}{c}{0.3855} & \multicolumn{1}{c||}{0.7186} & \multicolumn{1}{c}{0.2445} & \multicolumn{1}{c}{0.5878} & \multicolumn{1}{c}{0.4632} & \multicolumn{1}{c}{0.6688} & \multicolumn{1}{c}{0.8196} & \multicolumn{1}{c}{0.7731}\\
    
    \multicolumn{1}{c||}{Pima(35\%)} & \multicolumn{1}{c|}{0.4910} & \multicolumn{1}{c}{0.4937} & \multicolumn{1}{c}{0.4800} &   \multicolumn{1}{c}{0.4028} & \multicolumn{1}{c|}{0.5775} & \multicolumn{1}{c}{0.4983} & \multicolumn{1}{c||}{0.4983} & \multicolumn{1}{c}{0.3374} & \multicolumn{1}{c}{0.4419} & \multicolumn{1}{c}{0.5318} & \multicolumn{1}{c}{\textbf{0.6962}} & \multicolumn{1}{c}{0.6816} & \multicolumn{1}{c}{0.6852}\\
    
    \multicolumn{1}{c||}{Satellite (32\%)} & \multicolumn{1}{c|}{\textbf{0.6602}} & \multicolumn{1}{c}{0.5221} & \multicolumn{1}{c}{0.5031} &   \multicolumn{1}{c}{N/A} & \multicolumn{1}{c|}{0.5913} & \multicolumn{1}{c}{0.5007} & \multicolumn{1}{c||}{0.5000} & \multicolumn{1}{c}{0.3152} & \multicolumn{1}{c}{0.5481} & \multicolumn{1}{c}{0.5448} & \multicolumn{1}{c}{0.5717} & \multicolumn{1}{c}{0.6069} & \multicolumn{1}{c}{0.5997}\\
    
    \multicolumn{1}{c||}{Satimage-2 (1.2\%)} & \multicolumn{1}{c|}{0.8192} & \multicolumn{1}{c}{0.5903} & \multicolumn{1}{c}{0.5513} &   \multicolumn{1}{c}{0.1552} & \multicolumn{1}{c|}{0.7921} & \multicolumn{1}{c}{0.5002} & \multicolumn{1}{c||}{0.5000} & \multicolumn{1}{c}{0.9351} & \multicolumn{1}{c}{0.7048} & \multicolumn{1}{c}{0.5346} & \multicolumn{1}{c}{0.9707} & \multicolumn{1}{c}{\textbf{0.9925}} & \multicolumn{1}{c}{0.9879}\\
    
    \multicolumn{1}{c||}{Shuttle* (7\%)} & \multicolumn{1}{c|}{\textbf{0.9737}} & \multicolumn{1}{c}{0.6173} & \multicolumn{1}{c}{0.6051} &   \multicolumn{1}{c}{N/A} & \multicolumn{1}{c|}{0.6199} & \multicolumn{1}{c}{0.5007} & \multicolumn{1}{c||}{0.5007} & \multicolumn{1}{c}{0.1650} & \multicolumn{1}{c}{0.4528} & \multicolumn{1}{c}{0.6191} & \multicolumn{1}{c}{0.5859} & \multicolumn{1}{c}{0.7082} & \multicolumn{1}{c}{0.6737}\\
    
    \multicolumn{1}{c||}{Speech (1.65\%)} & \multicolumn{1}{c|}{\textbf{0.6435}} & \multicolumn{1}{c}{0.5492} & \multicolumn{1}{c}{0.6000} &   \multicolumn{1}{c}{0.2822} & \multicolumn{1}{c|}{0.5270} & \multicolumn{1}{c}{0.4494} & \multicolumn{1}{c||}{0.4513} & \multicolumn{1}{c}{0.5391} & \multicolumn{1}{c}{0.5162} & \multicolumn{1}{c}{0.5072} & \multicolumn{1}{c}{0.4791} & \multicolumn{1}{c}{0.4984} & \multicolumn{1}{c}{0.4909}\\
    
    \multicolumn{1}{c||}{Vertebral (12.5\%)} & \multicolumn{1}{c|}{0.2084} & \multicolumn{1}{c}{0.4906} & \multicolumn{1}{c}{0.5078} &   \multicolumn{1}{c}{\textbf{0.6390}} & \multicolumn{1}{c|}{0.4273} & \multicolumn{1}{c}{0.5762} & \multicolumn{1}{c||}{0.5786} & \multicolumn{1}{c}{0.6232} & \multicolumn{1}{c}{0.4127} & \multicolumn{1}{c}{0.5019} & \multicolumn{1}{c}{0.3222} & \multicolumn{1}{c}{0.3663} & \multicolumn{1}{c}{0.3575}\\
    
    \multicolumn{1}{c||}{Vowels (3.4\%)} & \multicolumn{1}{c|}{0.8802} & \multicolumn{1}{c}{{0.9467}} & \multicolumn{1}{c}{0.8716} &   \multicolumn{1}{c}{0.0168} & \multicolumn{1}{c|}{0.9074} & \multicolumn{1}{c}{0.0341} & \multicolumn{1}{c||}{0.0346} & \multicolumn{1}{c}{0.2929} & \multicolumn{1}{c}{0.8944} & \multicolumn{1}{c}{0.9399} & \multicolumn{1}{c}{0.9373} & \multicolumn{1}{c}{\textbf{0.9468}} & \multicolumn{1}{c}{0.9445}\\
    
    \multicolumn{1}{c||}{Wine (7.7\%)} & \multicolumn{1}{c|}{\textbf{0.9958}} & \multicolumn{1}{c}{0.9361} & \multicolumn{1}{c}{0.8899} &   \multicolumn{1}{c}{0.0185} & \multicolumn{1}{c|}{0.6765} & \multicolumn{1}{c}{0.5420} & \multicolumn{1}{c||}{0.5420} & \multicolumn{1}{c}{0.1857} & \multicolumn{1}{c}{0.9916} & \multicolumn{1}{c}{0.9899} & \multicolumn{1}{c}{0.8605} & \multicolumn{1}{c}{0.8739} & \multicolumn{1}{c}{0.9067}\\
    % \hhline{=||=|=|=|=|=||=|=|=|=|=|=}
    \hline
    \multicolumn{1}{c||}{\textbf{AVG}} & \multicolumn{1}{c|}{\textbf{0.7657}} & \multicolumn{1}{c}{0.6450} & \multicolumn{1}{c}{0.6318} &   \multicolumn{1}{c}{0.2680} & \multicolumn{1}{c|}{0.6841} & \multicolumn{1}{c}{0.5160} & \multicolumn{1}{c||}{0.5379} & \multicolumn{1}{c}{0.3266} & \multicolumn{1}{c}{0.5985} & \multicolumn{1}{c}{0.6414} & \multicolumn{1}{c}{0.7121} & \multicolumn{1}{c}{0.7403} & \multicolumn{1}{c}{0.7329}\\
    \hline
    \multicolumn{12}{l}{Note: (1) The performance of iPOF in this table is based on LOF as the first stage detector, (2) N/A indicates the out-of-memory error.  }\\
    \end{tabular}}\label{tab:performance}
\end{table*}

\subsection{Experimental Settings}
\textbf{Datasets}. To fully evaluate our iPOF algorithm, numerous data sets in different domains are employed. They include the hospital patients data such as \textit{BreastW}, \textit{Arrhythmia}, \textit{Cardio}, \textit{Pima}, and \textit{Vertebral}; handwritten digits data like \textit{Mnist}, \textit{Optdigits}, and \textit{Pendigits}; also satellite image data like \textit{Satellite} and \textit{Satimage-2}, and other multivariate data. These data sets are used in \textcolor{gray}{\cite{Aggarwal2015TheoreticalFA,Keller2012HiCSHC,LiuFei2008,bandaragoda2014efficient,SatheA16,micenkova2014learning}}, and they are all maintained by the ODDS~\textcolor{gray}{\cite{Rayana:2016}} library. Following the same experimental setting, we treat the class with smallest size as outliers. Table\textcolor{red}{~\ref{tab:datasets}} shows the numbers of instance, dimension, outlier, outlier ratio and the specific types of outlier class in these data sets.

\noindent\textbf{Competitive Methods} Here we choose three different kinds of outlier detectors. (1) Classical basic outlier detector methods. We choose a variety of methods including outlier detectors like Local Outlier Factor (LOF)~\textcolor{gray}{\cite{Breunig00SIR}}, Connectivity-based Outlier Factor (COF)~\textcolor{gray}{\cite{Tang02PKDD}}, Fast Angle-Based Outlier Detection (FABOD)~\textcolor{gray}{\cite{kriegel2008angle}}, Subspace Outlier Detection (SOD)~\textcolor{gray}{\cite{10.1007/978-3-642-01307-2_86}}. (2) Artificial outlier generation based methods. We include deep learning-based outlier detector like Multiple-Objective Generative Adversarial Active Learning (MGAAL)~\textcolor{gray}{\cite{liu2019generative}} and Single-Objective Generative Adversarial Active Learning (SGAAL)~\textcolor{gray}{\cite{liu2019generative}}. (3) Ensemble outlier detection methods. isolation Forest (iForest)~\textcolor{gray}{\cite{LiuFei2008}}, and outlier ensembles \& combination frameworks like Locally Selective Combination of Parallel Outlier Ensembles (LSCP)~\textcolor{gray}{\cite{DBLP:journals/corr/abs-1812-01528}}, Feature Bagging (FB)~\textcolor{gray}{\cite{Aggarwal2015TheoreticalFA}}, Combination of outlier detectors by taking the median of the scores (Median)~\textcolor{gray}{\cite{Aggarwal2015TheoreticalFA}}, Combination of outlier detectors by taking the Average of Maximum (AOM)~\textcolor{gray}{\cite{Aggarwal2015TheoreticalFA}} and Maximum of Average (MOA)~\textcolor{gray}{\cite{Aggarwal2015TheoreticalFA}} are included. The parameters of the above outlier detection methods are set as follows: the neighbor number is 10 in LOF, COF, FABOD, SOD; the sub-sampling size and tree number are 200 and 100 in iForest. For generative adversarial learning-based outlier detection MGAAL and SGAAL, we set the amount of contamination to 0.1, number of training epoch to 20, the learning rate of the generator to 0.0001, the learning rate of the discriminator to 0.01, the decay rate for SGD to 1e-6, and momentum of SGD to 0.9. All outlier detection algorithms used in the paper are implemented in Python by PyOD~\textcolor{gray}{\cite{zhao2019pyod}} toolbox.

\noindent\textbf{Validation Metric}. The receiver operating characteristic (ROC) curve drawn on the outlier scores serves as the evaluation metric. In binary classification, the class prediction for each instance is often made based on a continuous random variable \textit{X}, which is a "score" computed for the instance. Given a threshold parameter \textit{T}, the instance is classified as "positive" if \textit{X}$>$\textit{T}, and "negative" otherwise. \textit{X} follows a probability density $f_{1}(x)$ if the instance actually belongs to class "positive", and $f_{0}(x)$ if otherwise. Therefore, the true positive rate is given by ${\mbox{TPR}(T)=\int _{T}^{\infty }f_{1}(x)\,dx}$ and the false positive rate is given by ${\mbox{FPR}(T)=\int _{T}^{\infty }f_{0}(x)\,dx}$. The ROC curve plots parametrically TPR(\textit{T}) versus FPR(\textit{T}) with \textit{T} as the varying parameter. The area under the curve (AUC) is equal to the probability that a classifier will rank a randomly chosen positive instance higher than a randomly chosen negative one. Hence we use the area under the ROC curve (AUC) as accuracy metric ranging from 0 to 1, where the value 1 corresponds to a perfect outlier detection result. With given true-positive rate (TPR) and false-positive rate (FPR), Area Under Curve (AUC) can be computed as follows:
\begin{equation}
\begin{split}
    AUC &= \int_{x=0}^{1}(TPR(FPR^{-1}(x)))dx = \int_{\infty}^{-\infty}TPR(T)FPR^{'}(T)dT\\
      &=  \int_{-\infty}^{\infty}\int_{-\infty}^{\infty} I(T^{'}>T)f_{1}(T^{'})f_{0}(T)dT^{'}dT = P(X_{1} > X_{0}),
\end{split}    
\end{equation}where $X_{1}$ is the score for a positive instance and $X_{0}$ is the score for a negative instance, and $f_{0}$ and $f_{1}$ are probability densities.

\noindent\textbf{Environment.} All experiments were run on a PC with an Intel(R) Core(TM) i7-8750H CPU @ 2.20GHz and a 16 GB DDR4 RAM.

\begin{figure*}[t]
\centering
\resizebox{1\textwidth}{!}{
\begin{minipage}[b]{0.48\textwidth}
\subfigure[\textit{Glass}]{
    \includegraphics[width=0.48\textwidth]{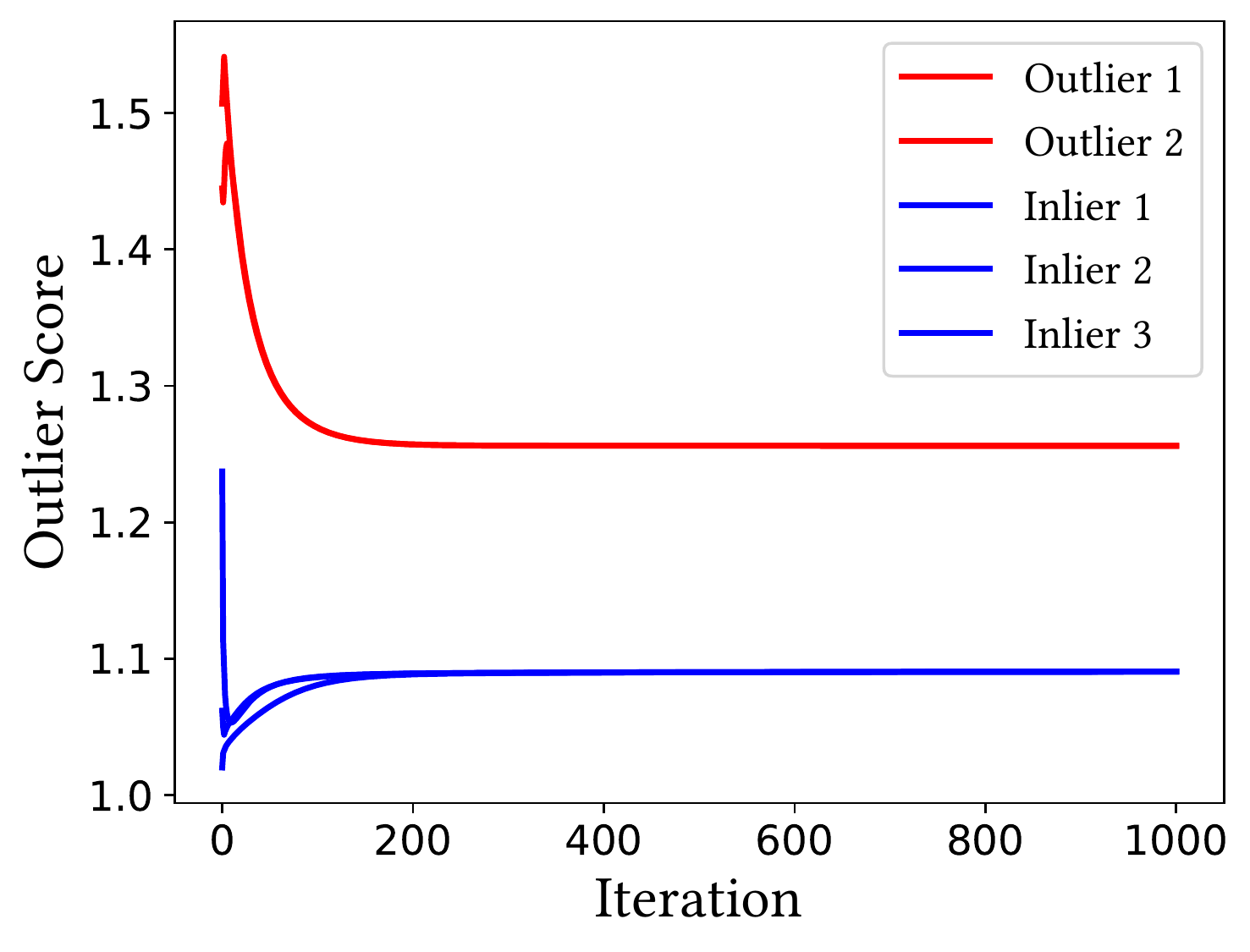}\label{fig:data-glass}}\hspace{-2mm}
    \subfigure[\textit{Mammography}]{
    \includegraphics[width=0.5\textwidth]{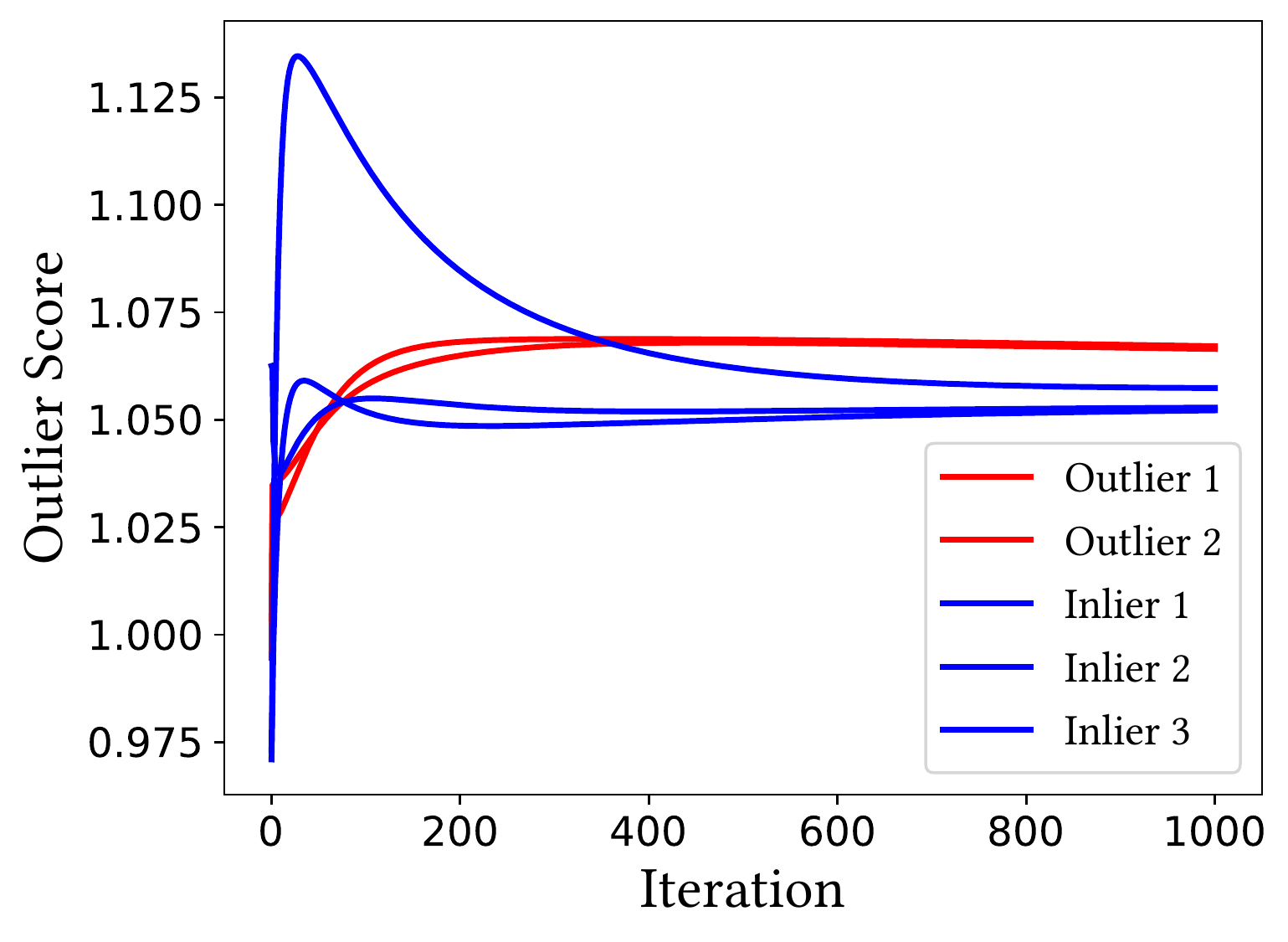}}

  \caption{Outlier scores via iPOF on \textit{Glass} and \textit{Mammography}, with tracks of several inliers and outliers that highlighted by blue and red color over iterations.}\label{fig:real_data}
\end{minipage}
\hspace{4mm}
\begin{minipage}[b]{0.48\textwidth}
\subfigure[\textit{Optdigits}]{
    \includegraphics[width=0.48\textwidth]{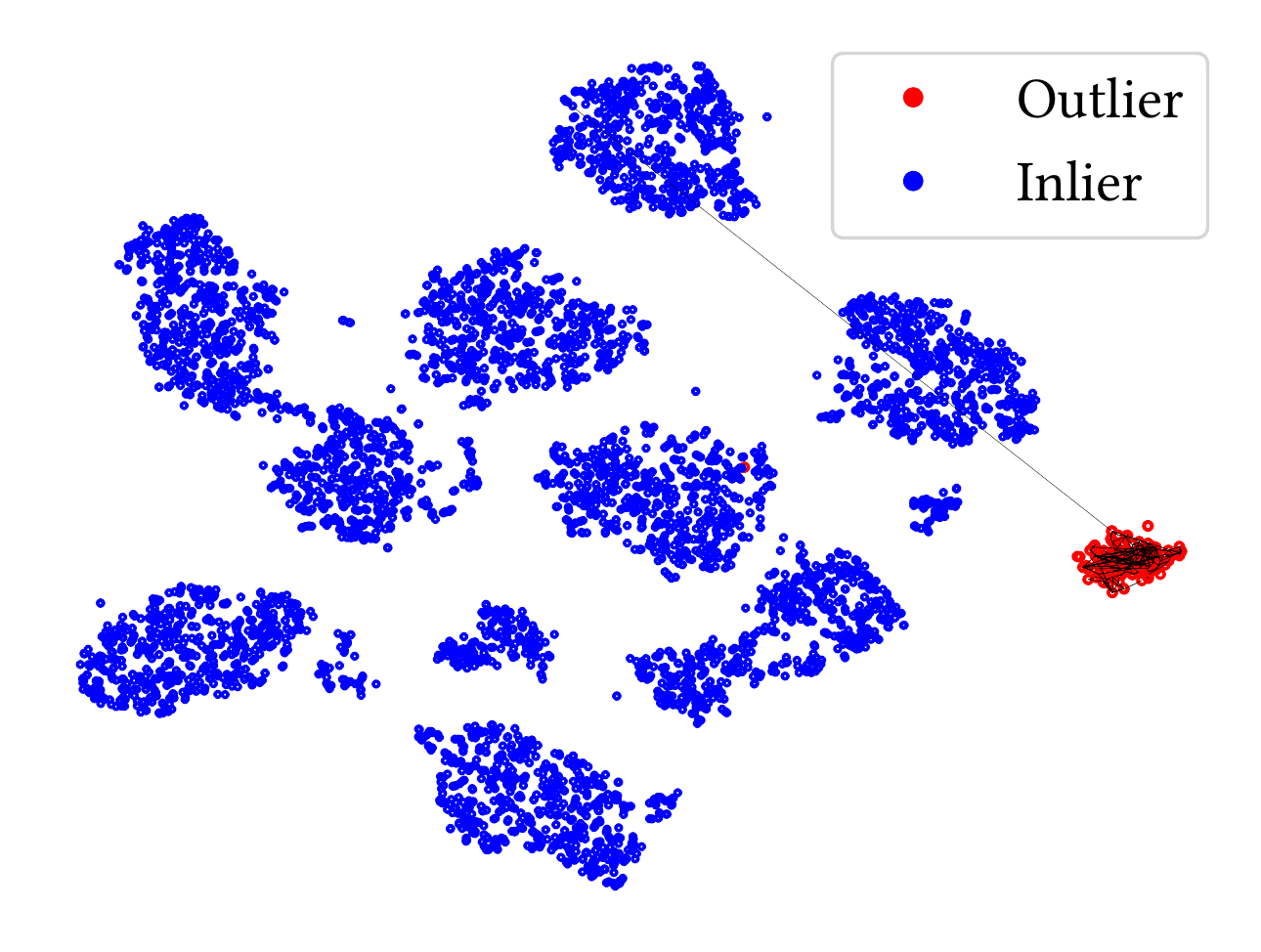}\label{fig:data-optdigits}}\hspace{-2mm}
    \subfigure[\textit{Vertebral}]{
    \includegraphics[width=0.48\textwidth]{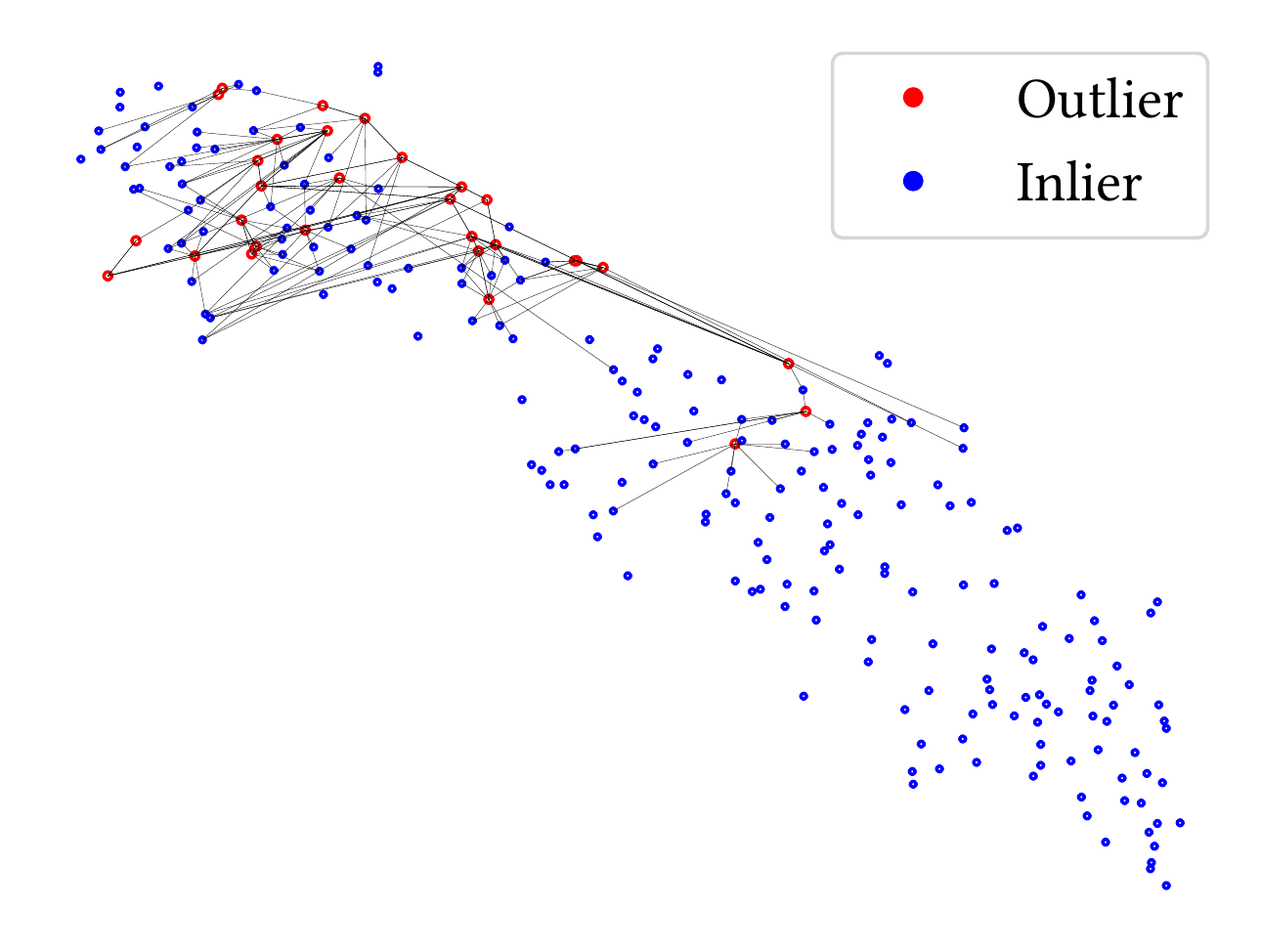}}

  \caption{Visualization of iPOF on \textit{Optdigits} (+57\% Improvement), \textit{Vertebral} (-57\% Improvement) by t-SNE\textcolor{gray}{\cite{maaten2008visualizing}}. Lines denote the neighborhood relationship of outliers.}\label{fig:edge_graph}
\end{minipage}}
\end{figure*}

\subsection{Algorithmic Performance}
\begin{figure}[t]
  \centering
    \subfigure[\textit{LOF}]{
    \includegraphics[width=0.237\textwidth]{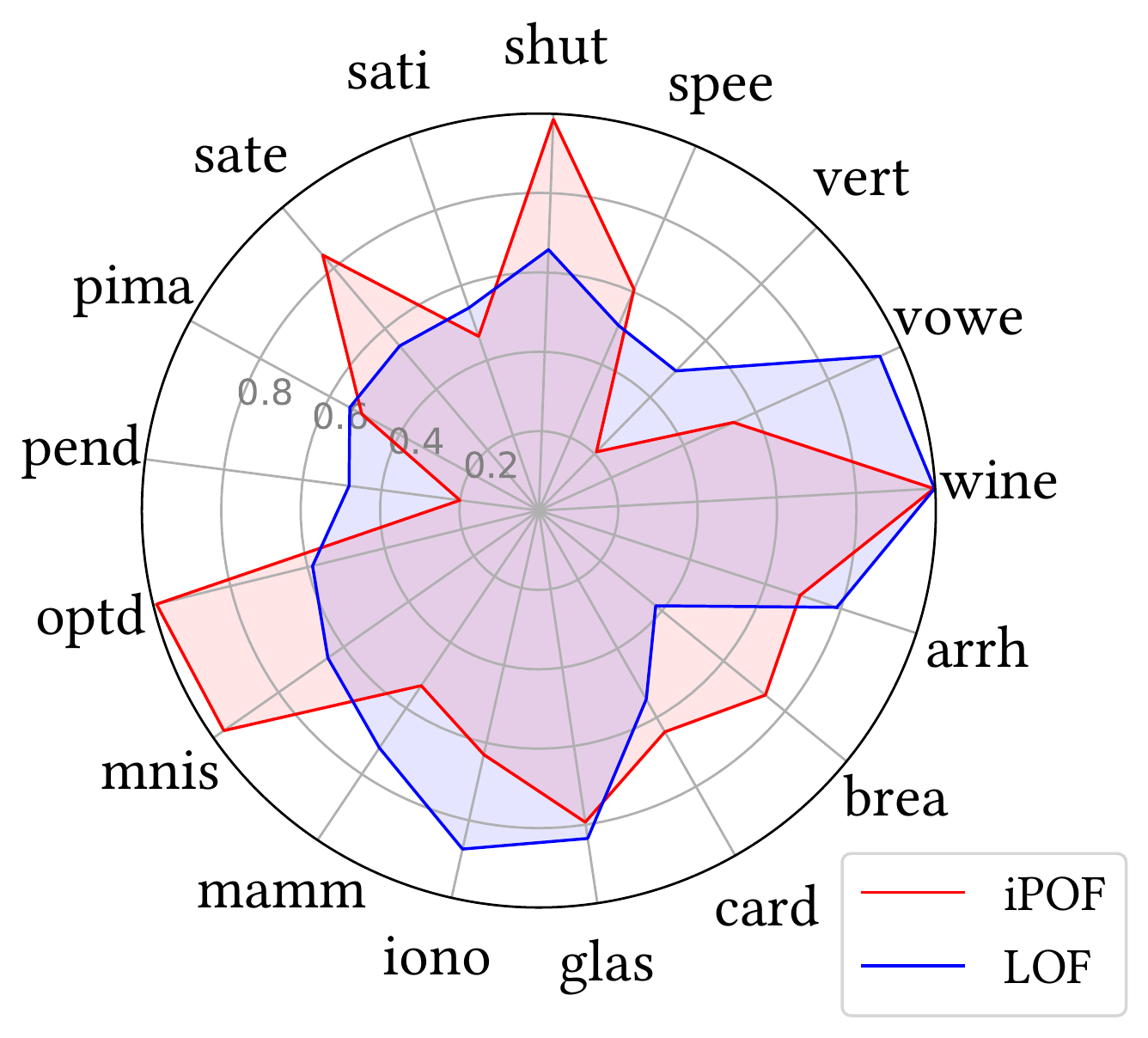}\label{fig:lof_20}}\hspace{-2mm}
    \subfigure[\textit{COF}]{
    \includegraphics[width=0.237\textwidth]{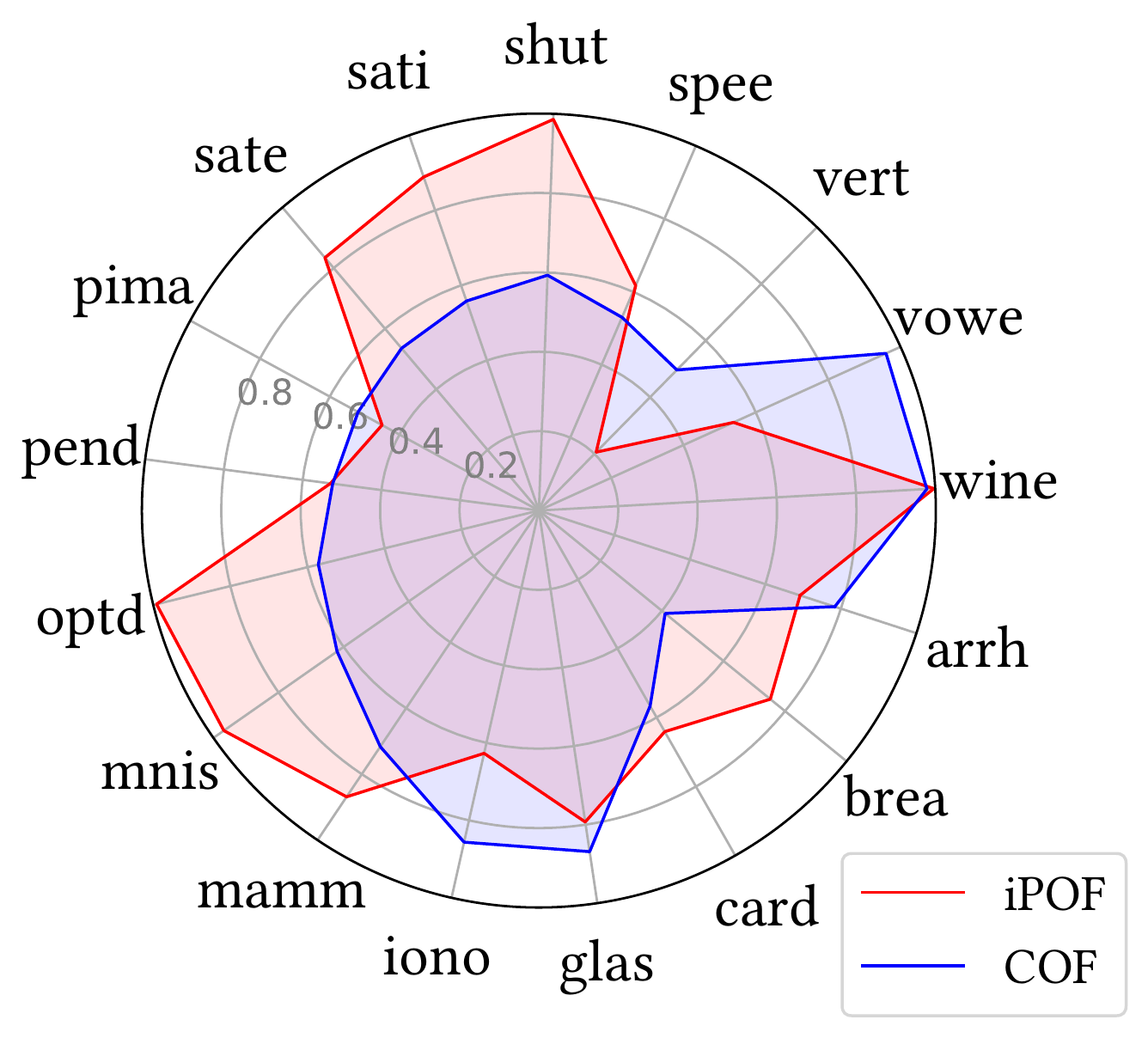}}\hspace{-6mm}
    \subfigure[\textit{FABOD}]{
    \includegraphics[width=0.237\textwidth]{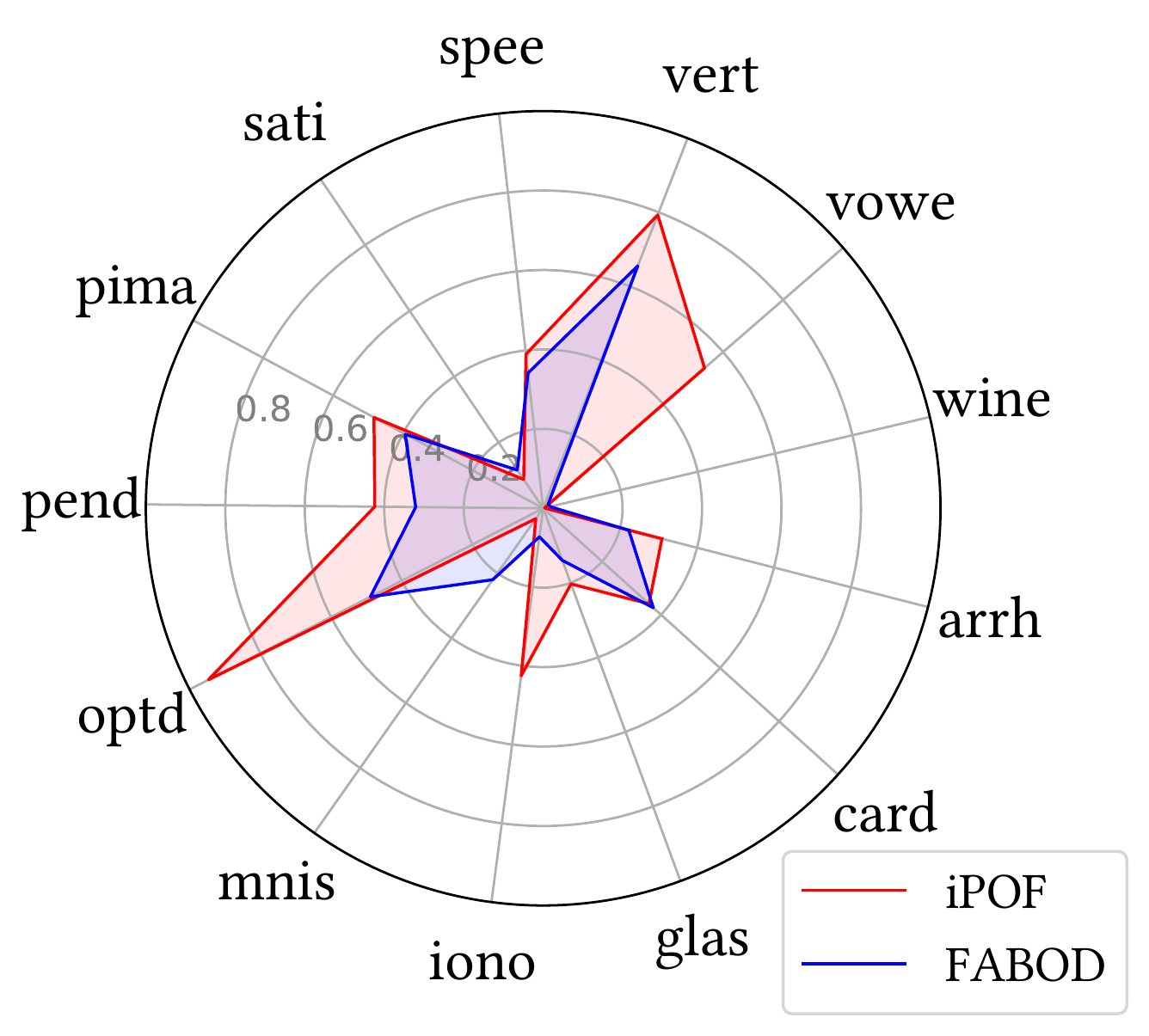}}\hspace{-2mm}
    \subfigure[\textit{SOD}]{
    \includegraphics[width=0.237\textwidth]{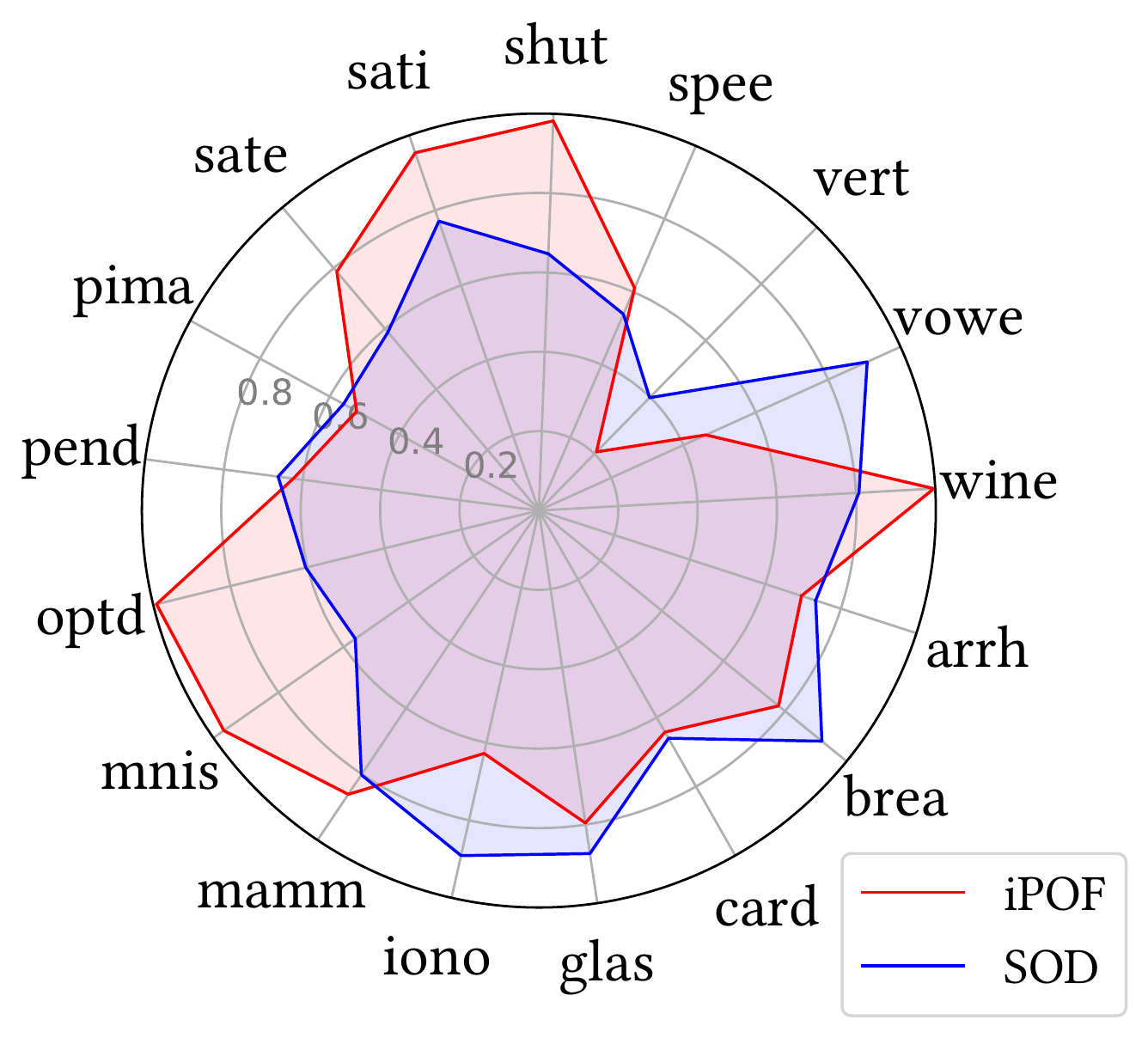}}
  \caption{AUC performance of iPOF on LOF, COF, FABOD and SOD as initialization with $K=20$. Each plot shows the performance of iPOF with particular detector on all data sets, where blue and orange area denote initial detector's performance and converged iPOF performance.} \label{fig:radar_plot}
\end{figure}

Table\textcolor{red}{~\ref{tab:performance}} shows the AUC performance of iPOF and other detectors on 17 datasets, where the best performance is highlighted in bold. We include 4 basic outlier detectors, 2 deep adversarial outlier detectors, and 6 ensemble outlier detectors. Note that iPOF works on the local neighborhood network with any initialized outlier scores. Here we only incorporate the basic outlier detector as the initialization, and the performance of iPOF in Table\textcolor{red}{~\ref{tab:performance}} is based on LOF. From the results in Table\textcolor{red}{~\ref{tab:performance}}, we elaborate several key observations. (1) Different outlier detectors achieve the best performance on different datasets. FABOD gets the best performance on \textit{Vertebral} by exceeding other basic outlier detectors by over 10\% in terms of AUC measurement; SOD, MGAAL, and SGAAL deliver the promising result on \textit{BreastW} and excels other outlier detectors by a large margin. The ensemble outlier detectors FB and Median achieve the best performance on \textit{Arrhythmia}, \textit{Ionosphere} and \textit{Pima}, respectively. Unfortunately, we can also see that FABOD returns extremely worst outlier candidates on \textit{Ionosphere}, \textit{Vowels} and \textit{Wine}. iForest cannot work well on \textit{BreastW}. These phenomena demonstrate that the assumptions of outlier algorithms are crucial to the success of effective detection. The unsupervised outlier detection setting also increases the task difficulty since we can only apply the default parameters. Usually, the ensemble outlier detectors outperform the basic outlier detectors due to their robustness by fusion mechanism that alleviates the dependence on algorithmic dependence. (2) Our iPOF algorithm achieves 9 out of 17 best results with the highest average performance as well. iPOF obtains 0.9676, 0.9562 and 0.9737 on \textit{Optdigits}, \textit{Pendigits} and \textit{Shuttle*}, that outperforms other basic and ensemble outlier detectors by around 30\%, 10\% and 30\%, respectively. Although iPOF also holds the assumption that inliers' core neighbors are inliers and outliers' core neighbors are outliers, the network built by common neighbors is robust and reliable in general. (3) Recall that the performance of iPOF in Table\textcolor{red}{~\ref{tab:performance}} is based on LOF as the initialization. If we take the comparisons between iPOF and LOF, there are significant improvements in 13 out of 17 datasets. Figure\textcolor{red}{~\ref{fig:lof_20}} provides a better visualization on the comparisons between iPOF and LOF on these 17 datasets with $K=20$. Therefore, LOF is not the termination point of outlier detection. iPOF starts from LOF and employs the local neighbor structure to further enhance the outlier detection performance just by simply averaging propagation. Admittedly, iPOF degrades the detection performance on some datasets. We conjecture that the parameter $K$ to choose the propagation neighbors plays a crucial role in whether there exist paths between outliers and inliers. In Table\textcolor{red}{~\ref{tab:performance}}, we set $K=10$ at the default. In the following subsection, we provide the parameter analysis and show the improvement on these datasets.

\begin{table*}[!t]
\caption {Running time of iPOF and other detectors by second} 

    \centering
    \resizebox{.98\textwidth}{!}{
    \begin{tabular}{c||cc|ccc|cc||ccccccc}
    \cline{1-15}
    % \hline
    % \multicolumn{1}{c||}{} & \multicolumn{2}{c|}{\textbf{Dataset}} \\
    \cline{2-14}
    \multicolumn{1}{c||}{\textbf{Datasets (size)}} & 
    \multicolumn{2}{c|}{\textbf{iPOF}}& 
    \multicolumn{1}{c}{\textbf{LOF}}& 
    \multicolumn{1}{c}{\textbf{COF}}& 
    \multicolumn{1}{c}{\textbf{FABOD}}&
    \multicolumn{1}{c|}{\textbf{SOD}}& 
    \multicolumn{1}{c}{\textbf{MGAAL}}&
    \multicolumn{1}{c||}{\textbf{SGAAL}}&
    \multicolumn{1}{c}{\textbf{iForest}}& 
    \multicolumn{1}{c}{\textbf{LSCP}}& 
    \multicolumn{1}{c}{\textbf{FB}}&  
    \multicolumn{1}{c}{\textbf{Median}}& 
    \multicolumn{1}{c}{\textbf{AOM}}& 
    \multicolumn{1}{c}{\textbf{MOA}}\\
    
    \hhline{=||==|====|==||======}
    \multicolumn{1}{c||}{Arrhythmia (452)} & \multicolumn{1}{c}{0.09} & 0.19& \multicolumn{1}{c}{0.10} &   \multicolumn{1}{c}{0.39} & \multicolumn{1}{c}{1.03} & \multicolumn{1}{c|}{3.93} & \multicolumn{1}{c}{10.43} & \multicolumn{1}{c||}{1.58} & \multicolumn{1}{c}{0.26} & \multicolumn{1}{c}{2.26} & \multicolumn{1}{c}{0.67} & \multicolumn{1}{c}{1.86} & \multicolumn{1}{c}{1.85} & \multicolumn{1}{c}{1.98}\\
    
    \multicolumn{1}{c||}{BreastW (683)} & \multicolumn{1}{c}{0.41} &0.42& \multicolumn{1}{c}{0.01} & \multicolumn{1}{c}{0.60} &   \multicolumn{1}{c}{N/A} & \multicolumn{1}{c|}{3.71} & \multicolumn{1}{c}{8.35} & \multicolumn{1}{c||}{1.12} & \multicolumn{1}{c}{0.19} & \multicolumn{1}{c}{0.53} & \multicolumn{1}{c}{0.07} & \multicolumn{1}{c}{0.38} & \multicolumn{1}{c}{0.37} & \multicolumn{1}{c}{0.40}\\
    
    \multicolumn{1}{c||}{Cardio (1831)} & \multicolumn{1}{c}{0.97}&1.07 & \multicolumn{1}{c}{0.10} & \multicolumn{1}{c}{3.94} &   \multicolumn{1}{c}{1.74} & \multicolumn{1}{c|}{5.09} & \multicolumn{1}{c}{15.68} & \multicolumn{1}{c||}{2.00} & \multicolumn{1}{c}{0.25} & \multicolumn{1}{c}{3.27} & \multicolumn{1}{c}{0.82} & \multicolumn{1}{c}{2.90} & \multicolumn{1}{c}{2.93} & \multicolumn{1}{c}{3.12}\\
    
    \multicolumn{1}{c||}{Glass (214)} & \multicolumn{1}{c}{0.08}&0.09 & \multicolumn{1}{c}{0.01} & \multicolumn{1}{c}{0.07} &   \multicolumn{1}{c}{0.73} & \multicolumn{1}{c|}{3.51} & \multicolumn{1}{c}{7.41} & \multicolumn{1}{c||}{0.86} & \multicolumn{1}{c}{0.17} & \multicolumn{1}{c}{0.15} & \multicolumn{1}{c}{0.03} & \multicolumn{1}{c}{0.07} & \multicolumn{1}{c}{0.07} & \multicolumn{1}{c}{0.07}\\
    
    \multicolumn{1}{c||}{Ionosphere (351)} & \multicolumn{1}{c}{0.20} &0.21& \multicolumn{1}{c}{0.01} & \multicolumn{1}{c}{0.17} &   \multicolumn{1}{c}{0.83} & \multicolumn{1}{c|}{3.59} & \multicolumn{1}{c}{8.53} & \multicolumn{1}{c||}{1.10} & \multicolumn{1}{c}{0.20} & \multicolumn{1}{c}{0.36} & \multicolumn{1}{c}{0.09} & \multicolumn{1}{c}{0.23} & \multicolumn{1}{c}{0.24} & \multicolumn{1}{c}{0.23}\\
    
    \multicolumn{1}{c||}{Mammography (11183)} & \multicolumn{1}{c}{7.95} & 8.26& \multicolumn{1}{c}{0.31} & \multicolumn{1}{c}{167.05} &   \multicolumn{1}{c}{N/A} & \multicolumn{1}{c|}{59.38} & \multicolumn{1}{c}{158.64} & \multicolumn{1}{c||}{12.27} & \multicolumn{1}{c}{0.56} & \multicolumn{1}{c}{N/A} & \multicolumn{1}{c}{2.91} & \multicolumn{1}{c}{11.05} & \multicolumn{1}{c}{11.51} & \multicolumn{1}{c}{11.35}\\
    
    \multicolumn{1}{c||}{Mnist (7603)} & \multicolumn{1}{c}{4.60} &14.65 & \multicolumn{1}{c}{10.05} & \multicolumn{1}{c}{105.25} &   \multicolumn{1}{c}{14.71} & \multicolumn{1}{c|}{37.24} & \multicolumn{1}{c}{120.66} & \multicolumn{1}{c||}{10.67} & \multicolumn{1}{c}{1.17} & \multicolumn{1}{c}{217.69} & \multicolumn{1}{c}{82.40} & \multicolumn{1}{c}{177.34} & \multicolumn{1}{c}{201.58} & \multicolumn{1}{c}{203.69}\\
    
    \multicolumn{1}{c||}{Optdigits (5126)} & \multicolumn{1}{c}{2.32} & 5.01 & \multicolumn{1}{c}{2.69} & \multicolumn{1}{c}{47.58} &   \multicolumn{1}{c}{6.26} & \multicolumn{1}{c|}{16.94} & \multicolumn{1}{c}{67.63} & \multicolumn{1}{c||}{6.65} & \multicolumn{1}{c}{0.65} & \multicolumn{1}{c}{63.43} & \multicolumn{1}{c}{20.74} & \multicolumn{1}{c}{64.57} & \multicolumn{1}{c}{68.13} & \multicolumn{1}{c}{69.05}\\
    
    \multicolumn{1}{c||}{Pendigits (6870)} & \multicolumn{1}{c}{6.48} & 6.87& \multicolumn{1}{c}{0.39} & \multicolumn{1}{c}{60.96} &   \multicolumn{1}{c}{4.47} & \multicolumn{1}{c|}{23.42} & \multicolumn{1}{c}{95.84} & \multicolumn{1}{c||}{7.39} & \multicolumn{1}{c}{0.51} & \multicolumn{1}{c}{13.04} & \multicolumn{1}{c}{2.41} & \multicolumn{1}{c}{19.22} & \multicolumn{1}{c}{20.81} & \multicolumn{1}{c}{21.24}\\
    
    \multicolumn{1}{c||}{Pima (768)} & \multicolumn{1}{c}{0.27} & 0.28 & \multicolumn{1}{c}{0.01} & \multicolumn{1}{c}{0.67} &   \multicolumn{1}{c}{1.03} & \multicolumn{1}{c|}{3.83} & \multicolumn{1}{c}{11.31} & \multicolumn{1}{c||}{1.09} & \multicolumn{1}{c}{0.20} & \multicolumn{1}{c}{0.55} & \multicolumn{1}{c}{0.06} & \multicolumn{1}{c}{0.48} & \multicolumn{1}{c}{0.54} & \multicolumn{1}{c}{0.51}\\
    
    \multicolumn{1}{c||}{Satellite (6435)} & \multicolumn{1}{c}{4.07} & 5.03& \multicolumn{1}{c}{0.96} & \multicolumn{1}{c}{57.74} &   \multicolumn{1}{c}{N/A} & \multicolumn{1}{c|}{21.57} & \multicolumn{1}{c}{104.96} & \multicolumn{1}{c||}{7.33} & \multicolumn{1}{c}{0.50} & \multicolumn{1}{c}{24.35} & \multicolumn{1}{c}{6.73} & \multicolumn{1}{c}{24.67} & \multicolumn{1}{c}{30.12} & \multicolumn{1}{c}{26.96}\\
    
    \multicolumn{1}{c||}{Satimage-2 (214)} & \multicolumn{1}{c}{3.29} & 4.13 &\multicolumn{1}{c}{0.84} & \multicolumn{1}{c}{48.25} &   \multicolumn{1}{c}{4.52} & \multicolumn{1}{c|}{18.27} & \multicolumn{1}{c}{78.42} & \multicolumn{1}{c||}{6.19} & \multicolumn{1}{c}{0.52} & \multicolumn{1}{c}{22.49} & \multicolumn{1}{c}{6.75} & \multicolumn{1}{c}{22.97} & \multicolumn{1}{c}{27.56} & \multicolumn{1}{c}{24.17}\\
    
    \multicolumn{1}{c||}{Shuttle* (4909)} & \multicolumn{1}{c}{5.92}& 6.10 & \multicolumn{1}{c}{0.18} & \multicolumn{1}{c}{28.15} &   \multicolumn{1}{c}{N/A} & \multicolumn{1}{c|}{13.56} & \multicolumn{1}{c}{54.89} & \multicolumn{1}{c||}{4.92} & \multicolumn{1}{c}{0.37} & \multicolumn{1}{c}{7.06} & \multicolumn{1}{c}{1.25} & \multicolumn{1}{c}{8.81} & \multicolumn{1}{c}{10.55} & \multicolumn{1}{c}{9.46}\\
     
    \multicolumn{1}{c||}{Speech (3686)} & \multicolumn{1}{c}{1.81} & 11.47 & \multicolumn{1}{c}{9.66} & \multicolumn{1}{c}{48.67} &   \multicolumn{1}{c}{13.19} & \multicolumn{1}{c|}{18.76} & \multicolumn{1}{c}{73.61} & \multicolumn{1}{c||}{12.33} & \multicolumn{1}{c}{1.73} & \multicolumn{1}{c}{227.41} & \multicolumn{1}{c}{76.60} & \multicolumn{1}{c}{211.37} & \multicolumn{1}{c}{227.59} & \multicolumn{1}{c}{217.55}\\
    
    \multicolumn{1}{c||}{Vertebral (240)} & \multicolumn{1}{c}{0.07} &0.08 & \multicolumn{1}{c}{0.01} & \multicolumn{1}{c}{0.08} &   \multicolumn{1}{c}{0.81} & \multicolumn{1}{c|}{3.43} & \multicolumn{1}{c}{8.49} & \multicolumn{1}{c||}{1.20} & \multicolumn{1}{c}{0.17} & \multicolumn{1}{c}{0.15} & \multicolumn{1}{c}{0.03} & \multicolumn{1}{c}{0.07} & \multicolumn{1}{c}{0.08} & \multicolumn{1}{c}{0.08}\\
    
    \multicolumn{1}{c||}{Vowels (1456)} & \multicolumn{1}{c}{0.54} & 0.57 & \multicolumn{1}{c}{0.03} & \multicolumn{1}{c}{2.43} &   \multicolumn{1}{c}{1.41} & \multicolumn{1}{c|}{4.33} & \multicolumn{1}{c}{14.20} & \multicolumn{1}{c||}{1.56} & \multicolumn{1}{c}{0.22} & \multicolumn{1}{c}{1.68} & \multicolumn{1}{c}{0.25} & \multicolumn{1}{c}{1.46} & \multicolumn{1}{c}{1.54} & \multicolumn{1}{c}{1.54}\\
    
    \multicolumn{1}{c||}{Wine (129)} & \multicolumn{1}{c}{0.03} &0.04 & \multicolumn{1}{c}{0.01} & \multicolumn{1}{c}{0.04} &   \multicolumn{1}{c}{0.74} & \multicolumn{1}{c|}{3.62} & \multicolumn{1}{c}{7.48} & \multicolumn{1}{c||}{0.92} & \multicolumn{1}{c}{0.16} & \multicolumn{1}{c}{0.09} & \multicolumn{1}{c}{0.02} & \multicolumn{1}{c}{0.04} & \multicolumn{1}{c}{0.04} & \multicolumn{1}{c}{0.04}\\
    % \hhline{=||=|=|=|=|=||=|=|=|=|=|=}
    \hline
    \multicolumn{1}{c||}{\textbf{AVG}} & \multicolumn{1}{c}{2.30}& 3.79 & \multicolumn{1}{c}{1.49} & \multicolumn{1}{c}{33.65} &   \multicolumn{1}{c}{3.96} & \multicolumn{1}{c|}{14.36} & \multicolumn{1}{c}{49.80} & \multicolumn{1}{c||}{4.66} & \multicolumn{1}{c}{0.46} & \multicolumn{1}{c}{36.53} & \multicolumn{1}{c}{11.87} & \multicolumn{1}{c}{32.20} & \multicolumn{1}{c}{35.62} & \multicolumn{1}{c}{34.79}\\
    \hline
    \multicolumn{15}{l}{Note: the first column denotes the score propagation of iPOF, while the second column is the total running time of iPOF with LOF as the initial detector.}\label{fig:time}\\
    \end{tabular}}
\end{table*}

\begin{figure*}[t]
  \centering
    \subfigure[\textit{Mnist}]{
    \includegraphics[width=0.34\textwidth]{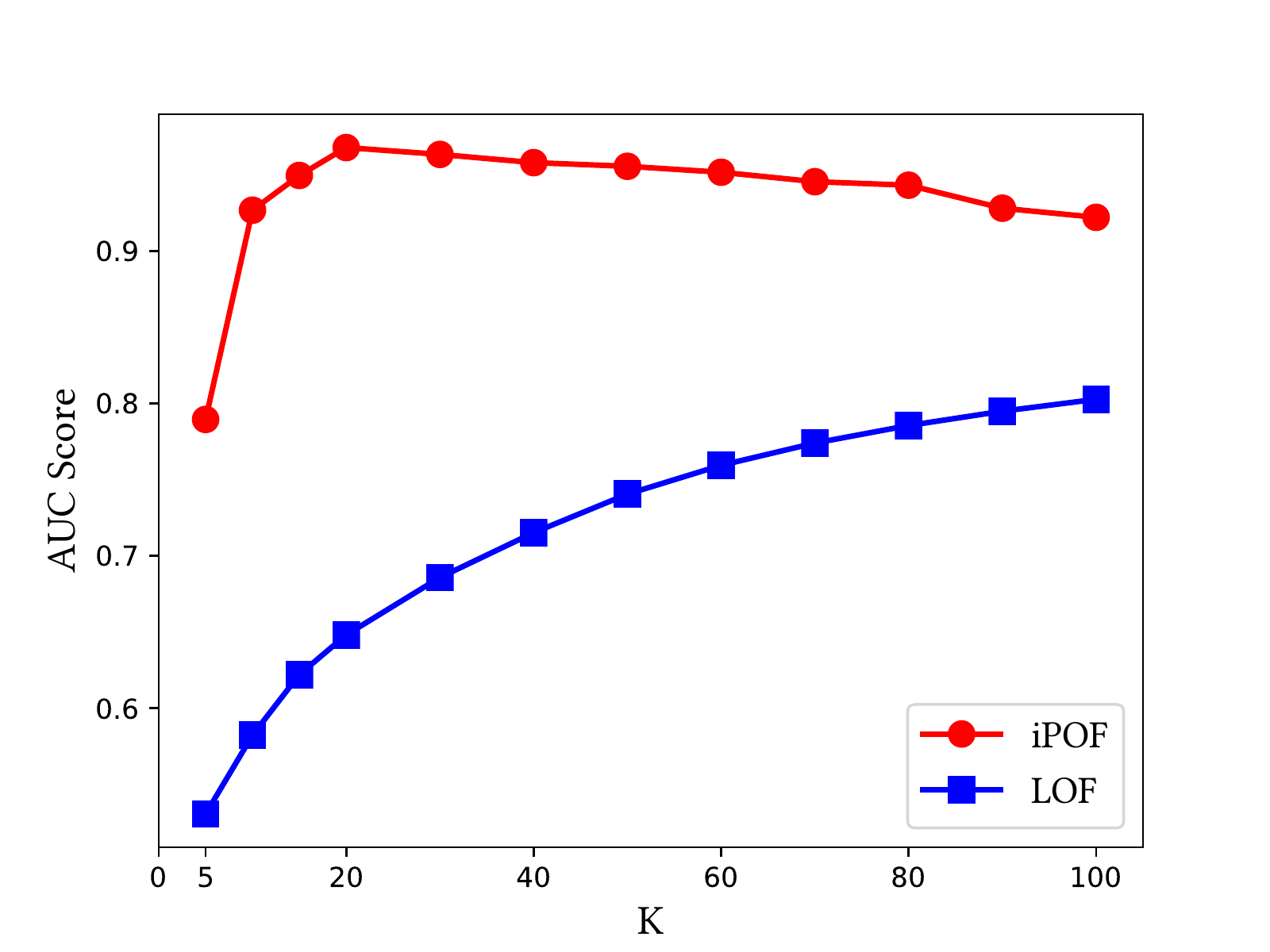}\label{fig:mnist-k-line}}\hspace{-6mm}
    \subfigure[\textit{Optdigits}]{
    \includegraphics[width=0.34\textwidth]{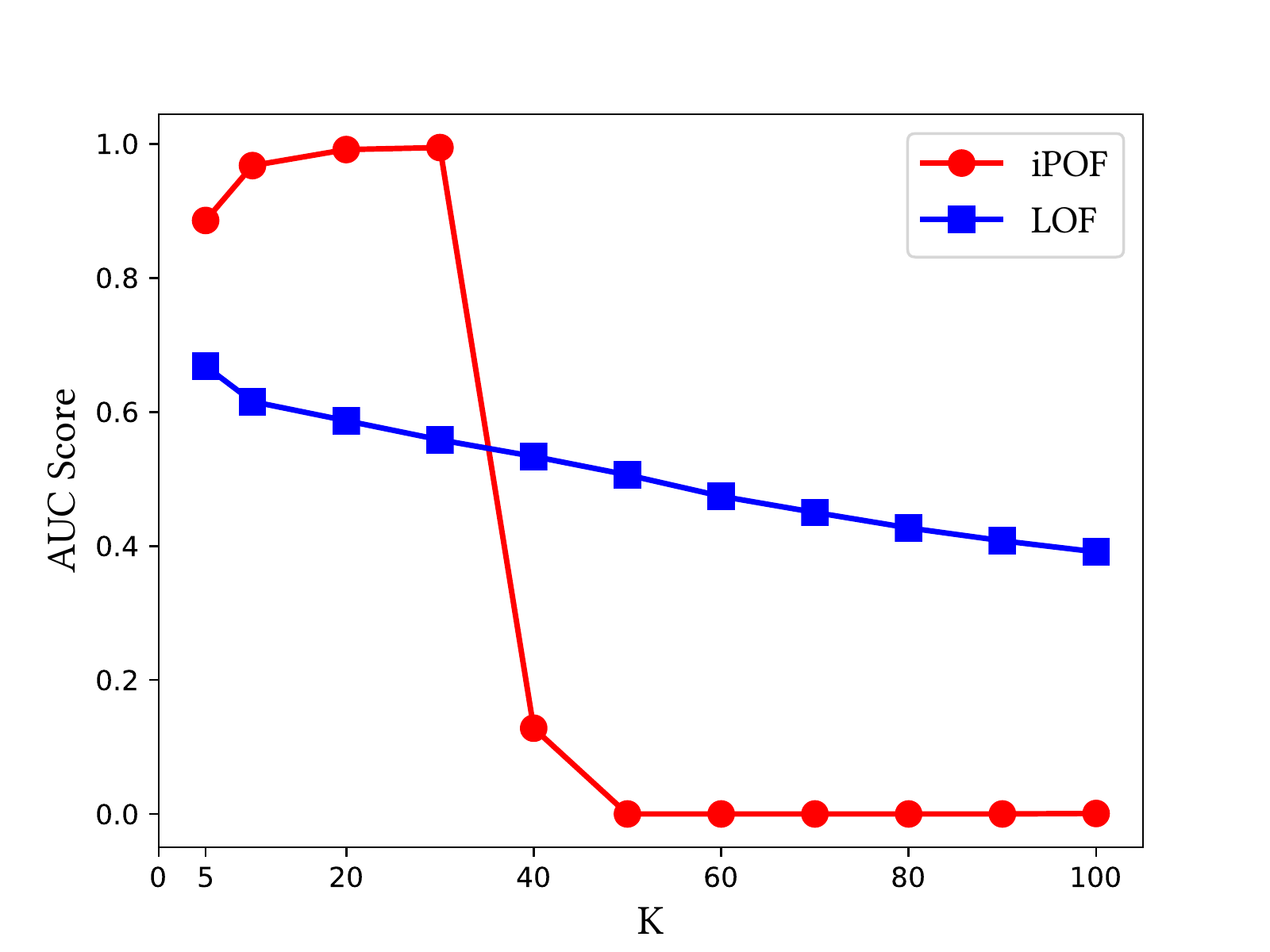}\label{fig:opt-k-line}}\hspace{-6mm}
    \subfigure[\textit{Speech}]{
    \includegraphics[width=0.34\textwidth]{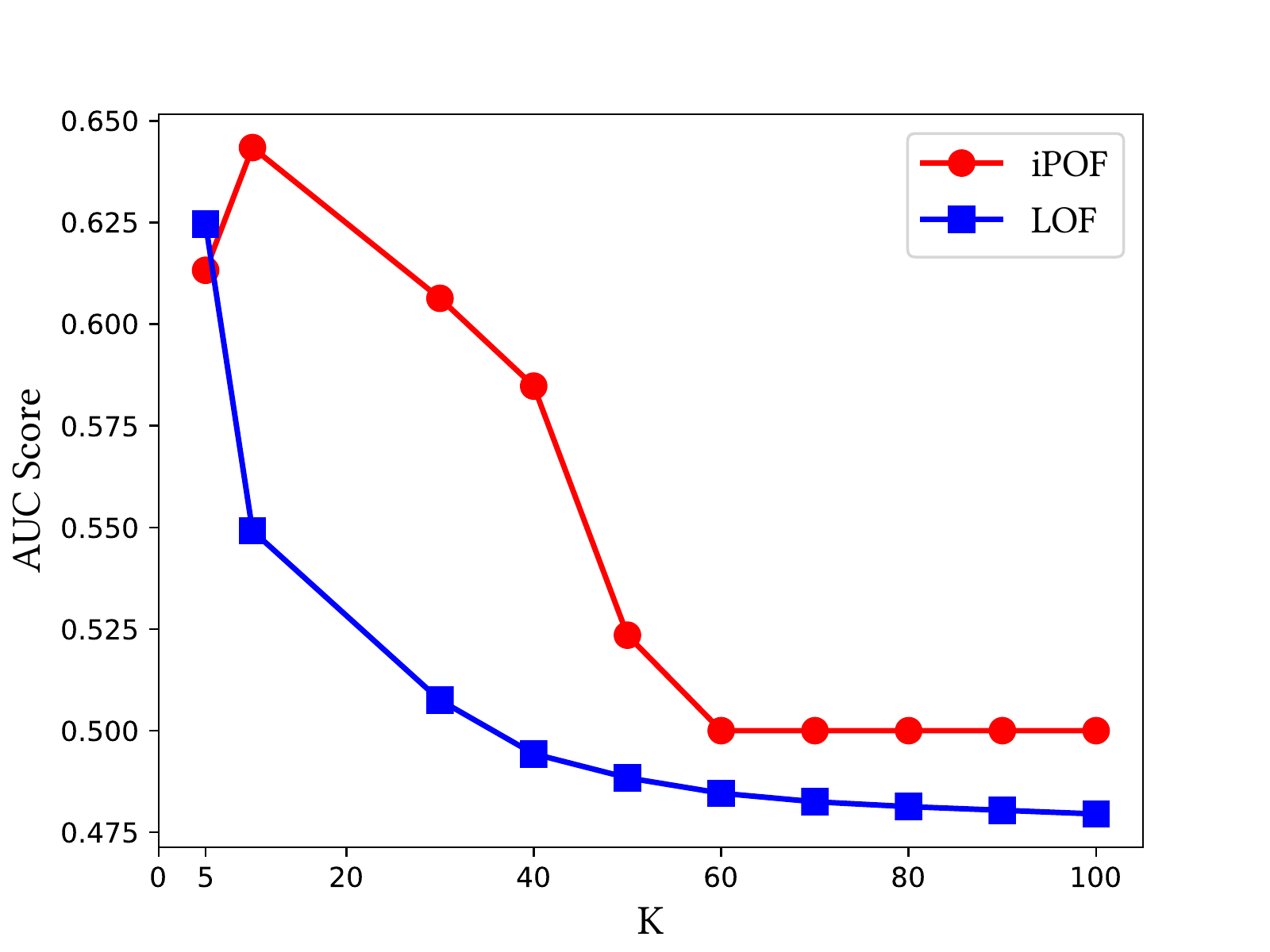}\label{fig:speech-k-line}}

  \caption{Parameter analysis on $K$ (ranging from 5 to 100) in iPOF on 3 real-world data sets \textit{Mnist}, \textit{Optdigits} and \textit{Speech}, with tracks of initial LOF detector's performance and converged iPOF that highlighted by red and blue color.} \label{fig:K}
\end{figure*}

Figure\textcolor{red}{~\ref{fig:real_data}} shows the outlier score trend with respect to the iteration of iPOF on \textit{Glass} and \textit{Mammography}, where red and blue lines denote outliers and inliers, respectively. Generally speaking, all lines converge during the iterations of iPOF, which validates the convergence of iPOF. On \textit{Mammography}, the scores of outliers and inliers are mixed at the initial stage. It is appealing to see that the scores of outliers increase, and the scores of inliers decrease, and they are separated after iPOF converges. Since the core idea of iPOF is to propagate the scores via averaging the local common neighbors, it is also normal to have the scores of some outliers decrease as well. Figure\textcolor{red}{~\ref{fig:edge_graph}} demonstrates the local neighborhood relationship on \textit{Optdigits} and \textit{Vertebral}, where red and blue nodes represent inlier and outlier samples, respectively. For better visualization, only the neighborhood relationship of outliers are shown in  Figure\textcolor{red}{~\ref{fig:edge_graph}}. We can see that iPOF brings in +57\% and -57\% improvements on these two datasets. Based on the local homogeneity assumption of iPOF, the positive or negative boost results from the local neighbor structure. On \textit{Optdigits}, outliers' neighbors are outliers and inliers' neighbors are inliers; on the contrary, some neighbors of outlier samples in \textit{Vertebral} are inliers, which leads to the negative propagation. In the unsupervised scenarios, algorithmic assumptions are crucially important and directly determine the ultimate performance. We do not claim that iPOF is helpful to boost basic outlier detectors on all datasets. Fortunately, iPOF brings in the positive improvements on the average level to most data sets. We will provide more solid experiments in the next subsection of in-depth factor exploration.

Table\textcolor{red}{~\ref{fig:time}} shows the running time of iPOF and other detectors. Generally speaking, iPOF is simple and fast. The most time-consuming part of iPOF is to build the common neighbor network based on data features in the first stage, which is pre-calculated in some density- or distance-based outlier detectors and can be directly used in iPOF. Moreover, some techniques including sampling~\textcolor{gray}{\cite{dudani1976distance}}, clustering~\textcolor{gray}{\cite{brito1997connectivity}} can be applied to quickly calculate the neighbor matrix. The second propagation stage just iteratively averages the outlier scores of top-$K$ common neighbors. The time complexity of the propagation stage is $\mathcal{O}(nK)$, which can be further accelerated by parallel computing. It is worthy to note that the input of the second stage is an $n\times 1$ outlier score vector, rather than the original data matrix $n\times d$, where $d$ is the number of feature dimensions. Therefore, iPOF is also space efficient and has $\mathcal{O}(n)$ in space complexity, which is essential to store the outlier score for every data point.

\vspace{-2mm}

\subsection{Factor Exploration}
In this subsection, we provide an in-depth exploration of iPOF in terms of parameter analysis and initial outlier detectors.

\noindent\textit{\textbf{Parameter Analysis on $K$}}. In Figure\textcolor{red}{~\ref{fig:neighbor}}, we introduce $k$ to construct the common neighbor network and select $K$ to outlier score propagation in iPOF. In fact, $k$ is used to illustrate the concept of an common neighbor network. We can set $k=n$ to get a fully connected network. Instead, the number of neighbors for propagation $K$ is the key factor in determining the outlier score propagation. Therefore, there is only one parameter $K$ in our iPOF algorithm. We provide the parameter analysis on $K$ in iPOF on 3 real-world datasets \textit{Mnist}, \textit{Optdigits} and \textit{Speech} in Figure\textcolor{red}{~\ref{fig:K}}, where $K$ varies from 5 to 100 and the performance of LOF is added as a baseline. Generally speaking, $K$ plays a crucial role in outlier detection, especially in unsupervised settings. We can see that the performance of LOF increases with an increasing $K$ on \textit{Mnist}, while it decreases on \textit{Optdigits} and \textit{Speech}. The performance of iPOF goes up and then decreases with an increasing $K$. On \textit{Optdigits} and \textit{Speech}, there exist the significant performance drop when $K$ is larger than 40. These phenomena result from that a large $K$ involves the path between inliers and outliers and propagates the high scores of outliers to inliers. Based on the above experimental results, we recommend a small $K$ in iPOF, such as $K=10$ or $K=20$ for effective propagation.

\noindent\textit{\textbf{Different Outlier Detectors as Initialization}}. So far we use LOF as the default initialization of our iPOF in the above experiments, and demonstrate that iPOF, a simple averaging propagation technique, can improve LOF by 18.71\% on average AUC performance. In fact, iPOF is orthogonal to outlier detectors, that any score-based outlier detector can be used as initialization for score propagation. To demonstrate more experimental results, we set $K=20$ in iPOF on 4 different detectors LOF, COF, FABOD and SOD in Figure\textcolor{red}{~\ref{fig:radar_plot}}. In most cases, iPOF brings the positive improvements with different initial outlier detectors. Specifically, iPOF delivers dramatic improvements over LOF around 83\%, 50\%, 70\%, 55\% on \textit{BreastW}, \textit{Mnist}, \textit{Optdigits} and \textit{Satellite}, respectively. On COF, iPOF also enjoys the huge improvements on \textit{BreastW}, \textit{Mammography}, \textit{Mnist}, \textit{Optdigits}, \textit{Satellite}, \textit{Satimage-2} and \textit{Shuttle*} with the increasing ratios from 21\% to 83\%. Similarly, some improvements of iPOF over SOD on 7 datasets as well. Excitingly note that iPOF enhances FABOD on \textit{Vowels} from 0.0171 to 0.5382, bringing about 3047\% improvements. iPOF has a unique converged solution with a fixed common neighbor network, and different initialization leads to different final AUC scores. We can still see some negative propagation cases in Figure\textcolor{red}{~\ref{fig:radar_plot}}. The simple assumption of iPOF cannot capture all the scenarios of real-world datasets. Fortunately, iPOF works well on generally average level. Table\textcolor{red}{~\ref{tab:detector}} shows average AUC performance of iPOF with different initial outlier detectors on 17 datasets with $K=10$ and $K=20$, where iPOF brings the positive improvements ranging from 2\% to 46\%.

\begin{table}[t]
\caption {Average AUC performance of iPOF with different outlier detectors as initialization} \vspace{-0.3cm}
\resizebox{.48\textwidth}{!}{
\begin{tabular}{c|ccc|ccc}
\hline
\multirow{2}{*}{Detector}& \multicolumn{3}{c|}{$K=10$} & \multicolumn{3}{c}{$K=20$}\\
\cline{2-7}
 & Original & iPOF & Impr.\%  & Original & iPOF & Impr.\% \\
\hline \hline
LOF & 0.6450 & 0.7657 & 18.71 & 0.6327 & 0.6595 & 4.24\\
COF & 0.6318 & 0.6669 & 5.56 & 0.6523 & 0.7282 & 11.64\\ 
FABOD & 0.2680 & 0.3755 & 40.11 & 0.2592 & 0.3804 & 46.76\\
SOD & 0.6841 & 0.6984 & 2.09 & 0.7018 & 0.7357 & 4.83\\
\hline
\end{tabular}}\label{tab:detector}
\end{table}  

\vspace{-0mm}

\begin{figure}[t]
  \centering
    \subfigure[\textit{REPEN}]{
    \includegraphics[width=0.23\textwidth]{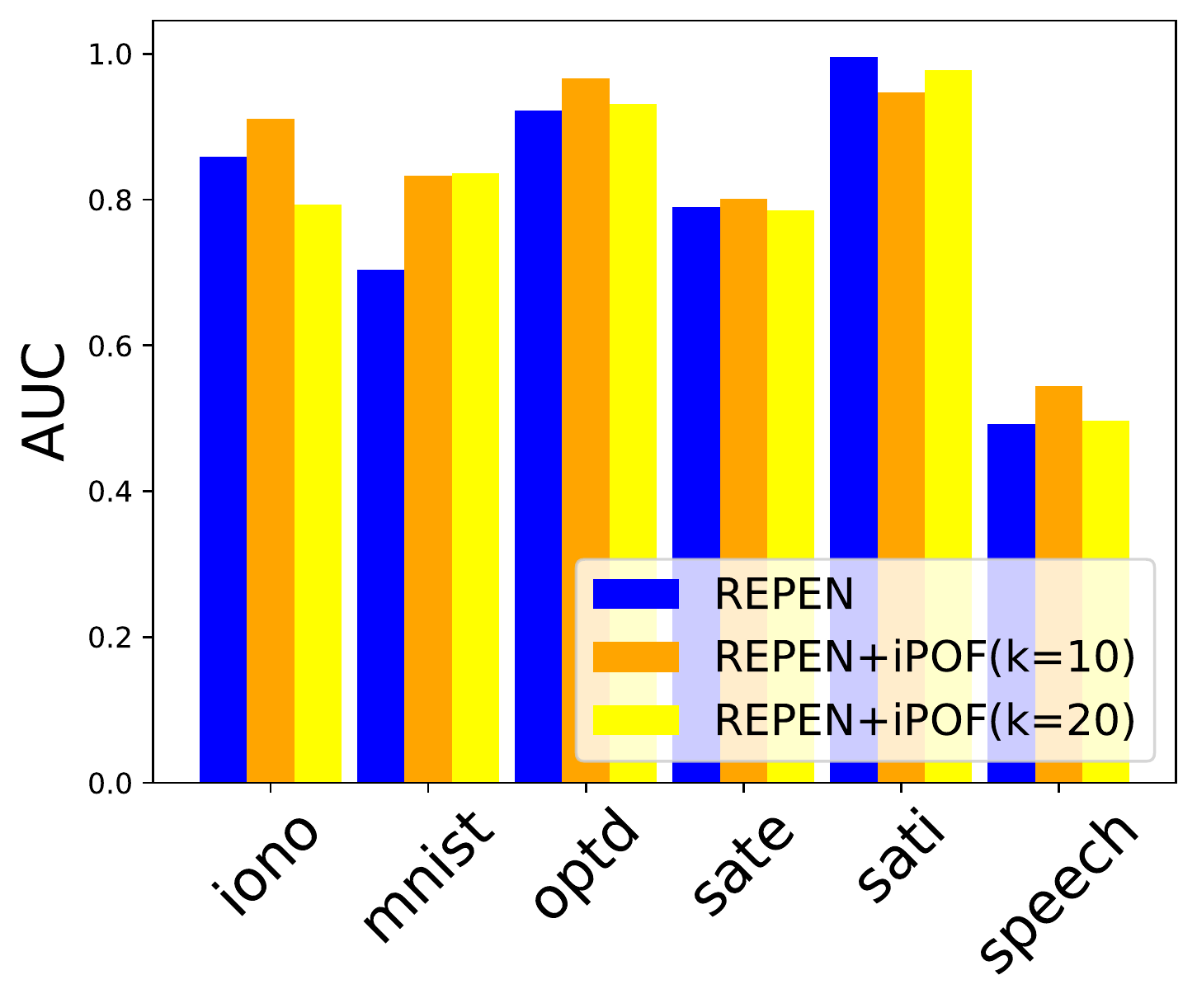}}
    \subfigure[\textit{RDP}]{
    \includegraphics[width=0.23\textwidth]{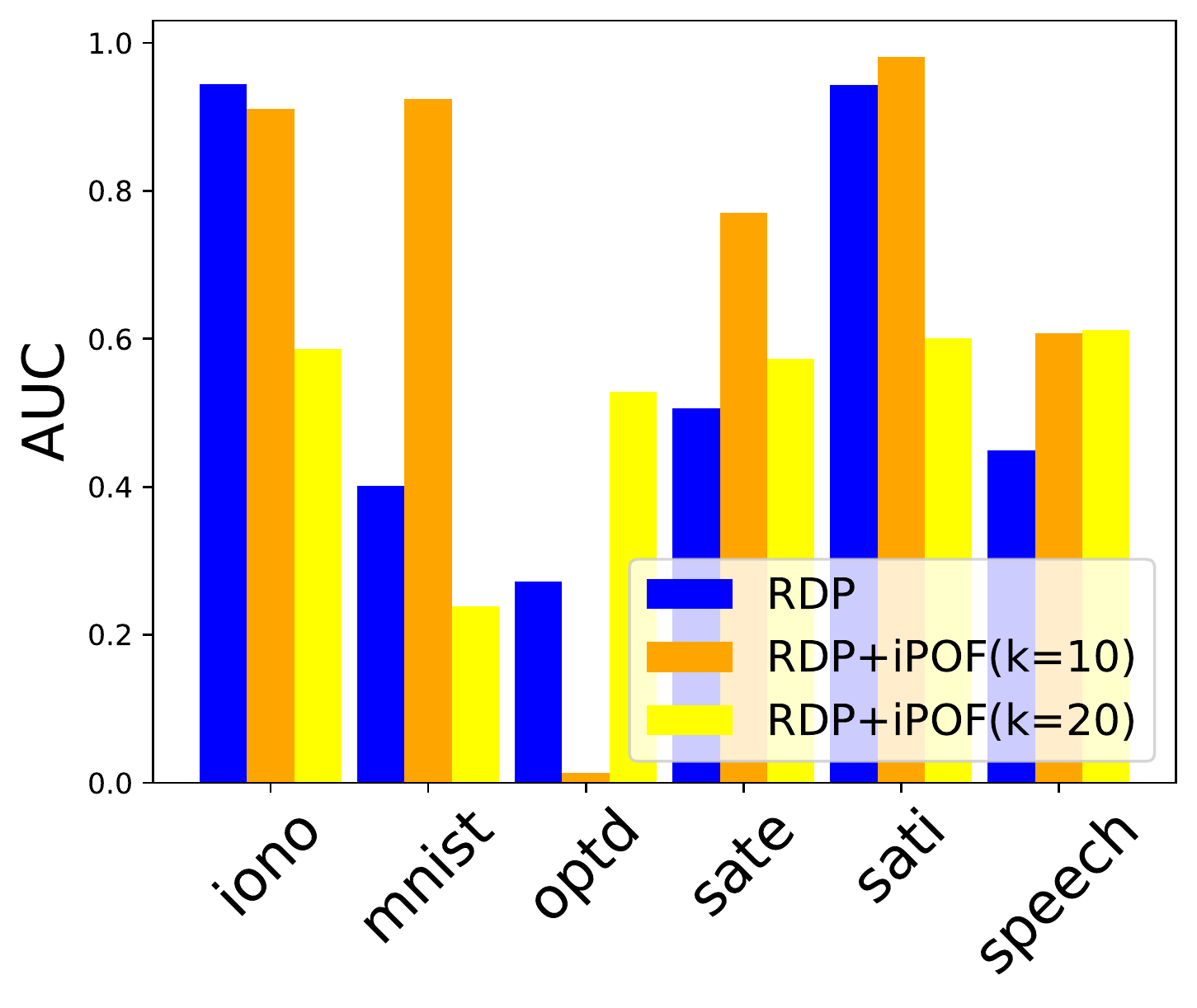}}
  \caption{iPOF performance on deep unsupervised outlier detection methods REPEN~\cite{pang2018learning} and RDP~\cite{wang2019unsupervised}.} \vspace{-4mm}\label{fig:deep}
\end{figure}

\noindent\textbf{iPOF on Deep Outlier Detection Methods}. As an outlier detection booster, we have demonstrated how iPOF helps on the conventional outlier detector methods so far. Actually, iPOF can be also applied on the deep outlier detection methods as well. Usually, the new representations are learned with the deep unsupervised outlier detection methods, where iPOF can build the common neighbor graph based on the learned representation, rather than the original feature space. Figure~\ref{fig:deep} shows the iPOF performance on recent deep unsupervised outlier detection methods REPEN~\cite{pang2018learning} and RDP~\cite{wang2019unsupervised} with different $K$ values. We can see that iPOF boosts the performance of RDP on \textit{glass}, \textit{mnist}, \textit{satellite} and \textit{satimage-2} by a large margin. However, the performance on \textit{optdigits} with $K=10$ drops a lot. This might result from the impact of the number of common neighbors. Recall that iPOF is an extremely simple method with only one parameter. 

\section{Conclusion}\label{sec:conclusion}
In this paper, we considered the unsupervised outlier detection problem and proposed the algorithm Infinite Propagation of Outlier Factor (iPOF), an extremely and excitingly simple outlier detector via infinite propagation. Under the assumption of local homogeneity that one data point has the neighbors of the same class, iPOF iteratively repeated the propagation process until convergence of outlier and inlier scores for better distinguishing. To achieve this, we first initialized iPOF with existing score-based outlier detectors. With awareness of neighborhood connectivity of each data point, iPOF then updated the outlier score of each instance by averaging the outlier factors of all its connected neighbors. Extensive experimental results demonstrated the effectiveness and efficiency of iPOF significantly over numerous classical and state-of-the-art unsupervised outlier detection and ensemble algorithms.

\bibliographystyle{ACM-Reference-Format}
\bibliography{ref}

%%% -*-BibTeX-*-
%%% Do NOT edit. File created by BibTeX with style
%%% ACM-Reference-Format-Journals [18-Jan-2012].

\begin{thebibliography}{55}

%%% ====================================================================
%%% NOTE TO THE USER: you can override these defaults by providing
%%% customized versions of any of these macros before the \bibliography
%%% command.  Each of them MUST provide its own final punctuation,
%%% except for \shownote{}, \showDOI{}, and \showURL{}.  The latter two
%%% do not use final punctuation, in order to avoid confusing it with
%%% the Web address.
%%%
%%% To suppress output of a particular field, define its macro to expand
%%% to an empty string, or better, \unskip, like this:
%%%
%%% \newcommand{\showDOI}[1]{\unskip}   % LaTeX syntax
%%%
%%% \def \showDOI #1{\unskip}           % plain TeX syntax
%%%
%%% ====================================================================

\ifx \showCODEN    \undefined \def \showCODEN     #1{\unskip}     \fi
\ifx \showDOI      \undefined \def \showDOI       #1{#1}\fi
\ifx \showISBNx    \undefined \def \showISBNx     #1{\unskip}     \fi
\ifx \showISBNxiii \undefined \def \showISBNxiii  #1{\unskip}     \fi
\ifx \showISSN     \undefined \def \showISSN      #1{\unskip}     \fi
\ifx \showLCCN     \undefined \def \showLCCN      #1{\unskip}     \fi
\ifx \shownote     \undefined \def \shownote      #1{#1}          \fi
\ifx \showarticletitle \undefined \def \showarticletitle #1{#1}   \fi
\ifx \showURL      \undefined \def \showURL       {\relax}        \fi
% The following commands are used for tagged output and should be
% invisible to TeX
\providecommand\bibfield[2]{#2}
\providecommand\bibinfo[2]{#2}
\providecommand\natexlab[1]{#1}
\providecommand\showeprint[2][]{arXiv:#2}

\bibitem[\protect\citeauthoryear{Aggarwal}{Aggarwal}{2013}]%
        {aggarwal2013outlier}
\bibfield{author}{\bibinfo{person}{Charu~C. Aggarwal}.}
  \bibinfo{year}{2013}\natexlab{}.
\newblock \showarticletitle{Outlier analysis: Springer science \& business
  media}.
\newblock  (\bibinfo{year}{2013}).
\newblock


\bibitem[\protect\citeauthoryear{Aggarwal and Sathe}{Aggarwal and
  Sathe}{2015}]%
        {Aggarwal2015TheoreticalFA}
\bibfield{author}{\bibinfo{person}{Charu~C. Aggarwal} {and}
  \bibinfo{person}{Saket Sathe}.} \bibinfo{year}{2015}\natexlab{}.
\newblock \showarticletitle{Theoretical foundations and algorithms for outlier
  ensembles}.
\newblock \bibinfo{journal}{\emph{SIGKDD Explorations}}  \bibinfo{volume}{17}
  (\bibinfo{year}{2015}), \bibinfo{pages}{24--47}.
\newblock


\bibitem[\protect\citeauthoryear{Agrawal and Agrawal}{Agrawal and
  Agrawal}{2015}]%
        {agrawal2015survey}
\bibfield{author}{\bibinfo{person}{Shikha Agrawal} {and}
  \bibinfo{person}{Jitendra Agrawal}.} \bibinfo{year}{2015}\natexlab{}.
\newblock \showarticletitle{Survey on anomaly detection using data mining
  techniques}.
\newblock \bibinfo{journal}{\emph{Procedia Computer Science}}
  \bibinfo{volume}{60} (\bibinfo{year}{2015}), \bibinfo{pages}{708--713}.
\newblock


\bibitem[\protect\citeauthoryear{Akcay, Atapour-Abarghouei, and Breckon}{Akcay
  et~al\mbox{.}}{2018}]%
        {akcay2018ganomaly}
\bibfield{author}{\bibinfo{person}{Samet Akcay}, \bibinfo{person}{Amir
  Atapour-Abarghouei}, {and} \bibinfo{person}{Toby~P Breckon}.}
  \bibinfo{year}{2018}\natexlab{}.
\newblock \showarticletitle{Ganomaly: Semi-supervised anomaly detection via
  adversarial training}. In \bibinfo{booktitle}{\emph{Proceedings of Asian
  Conference on Computer Vision}}. \bibinfo{pages}{622--637}.
\newblock


\bibitem[\protect\citeauthoryear{Bandaragoda, Ting, Albrecht, Liu, and
  Wells}{Bandaragoda et~al\mbox{.}}{2014}]%
        {bandaragoda2014efficient}
\bibfield{author}{\bibinfo{person}{Tharindu~R Bandaragoda},
  \bibinfo{person}{Kai~Ming Ting}, \bibinfo{person}{David Albrecht},
  \bibinfo{person}{Fei~Tony Liu}, {and} \bibinfo{person}{Jonathan~R Wells}.}
  \bibinfo{year}{2014}\natexlab{}.
\newblock \showarticletitle{Efficient anomaly detection by isolation using
  nearest neighbour ensemble}. In \bibinfo{booktitle}{\emph{Proceedings of IEEE
  International Conference on Data Mining Workshop}}.
  \bibinfo{pages}{698--705}.
\newblock


\bibitem[\protect\citeauthoryear{Breunig, Kriegel, Ng, and Sander}{Breunig
  et~al\mbox{.}}{2000}]%
        {Breunig00SIR}
\bibfield{author}{\bibinfo{person}{Markus~M Breunig},
  \bibinfo{person}{Hans-Peter Kriegel}, \bibinfo{person}{Raymond~T Ng}, {and}
  \bibinfo{person}{J{\"o}rg Sander}.} \bibinfo{year}{2000}\natexlab{}.
\newblock \showarticletitle{LOF: identifying density-based local outliers}. In
  \bibinfo{booktitle}{\emph{Proceedings of ACM SIGMOD record}},
  Vol.~\bibinfo{volume}{29}. \bibinfo{pages}{93--104}.
\newblock


\bibitem[\protect\citeauthoryear{Brito, Chavez, Quiroz, and Yukich}{Brito
  et~al\mbox{.}}{1997}]%
        {brito1997connectivity}
\bibfield{author}{\bibinfo{person}{María~Rosa Brito}, \bibinfo{person}{Edgar
  Chavez}, \bibinfo{person}{Adolfo~J. Quiroz}, {and} \bibinfo{person}{Joseph~E.
  Yukich}.} \bibinfo{year}{1997}\natexlab{}.
\newblock \showarticletitle{Connectivity of the mutual K-Nearest-Neighbor graph
  in clustering and outlier detection}.
\newblock \bibinfo{journal}{\emph{Statistics \& Probability Letters}}
  \bibinfo{volume}{35}, \bibinfo{number}{1} (\bibinfo{year}{1997}),
  \bibinfo{pages}{33--42}.
\newblock


\bibitem[\protect\citeauthoryear{Chalapathy and Chawla}{Chalapathy and
  Chawla}{2019}]%
        {chalapathy2019deep}
\bibfield{author}{\bibinfo{person}{Raghavendra Chalapathy} {and}
  \bibinfo{person}{Sanjay Chawla}.} \bibinfo{year}{2019}\natexlab{}.
\newblock \showarticletitle{Deep learning for anomaly detection: A survey}.
\newblock \bibinfo{journal}{\emph{arXiv preprint arXiv:1901.03407}}
  (\bibinfo{year}{2019}).
\newblock


\bibitem[\protect\citeauthoryear{Chalapathy, Menon, and Chawla}{Chalapathy
  et~al\mbox{.}}{2018}]%
        {chalapathy2018anomaly}
\bibfield{author}{\bibinfo{person}{Raghavendra Chalapathy},
  \bibinfo{person}{Aditya~Krishna Menon}, {and} \bibinfo{person}{Sanjay
  Chawla}.} \bibinfo{year}{2018}\natexlab{}.
\newblock \showarticletitle{Anomaly detection using one-class neural networks}.
\newblock \bibinfo{journal}{\emph{arXiv preprint arXiv:1802.06360}}
  (\bibinfo{year}{2018}).
\newblock


\bibitem[\protect\citeauthoryear{Chawla and Gionis}{Chawla and Gionis}{2013}]%
        {chawla2013k}
\bibfield{author}{\bibinfo{person}{Sanjay Chawla} {and}
  \bibinfo{person}{Aristides Gionis}.} \bibinfo{year}{2013}\natexlab{}.
\newblock \showarticletitle{K-means--: A unified approach to clustering and
  outlier detection}. In \bibinfo{booktitle}{\emph{Proceedings of SIAM
  International Conference on Data Mining}}.
\newblock


\bibitem[\protect\citeauthoryear{D{\'e}sir, Bernard, Petitjean, and
  Heutte}{D{\'e}sir et~al\mbox{.}}{2013}]%
        {desir2013one}
\bibfield{author}{\bibinfo{person}{Chesner D{\'e}sir}, \bibinfo{person}{Simon
  Bernard}, \bibinfo{person}{Caroline Petitjean}, {and}
  \bibinfo{person}{Laurent Heutte}.} \bibinfo{year}{2013}\natexlab{}.
\newblock \showarticletitle{One class random forests}.
\newblock \bibinfo{journal}{\emph{Pattern Recognition}} \bibinfo{volume}{46},
  \bibinfo{number}{12} (\bibinfo{year}{2013}), \bibinfo{pages}{3490--3506}.
\newblock


\bibitem[\protect\citeauthoryear{Dudani}{Dudani}{1976}]%
        {dudani1976distance}
\bibfield{author}{\bibinfo{person}{Sahibsingh~A Dudani}.}
  \bibinfo{year}{1976}\natexlab{}.
\newblock \showarticletitle{The distance-weighted K-Nearest-Neighbor rule}.
\newblock \bibinfo{journal}{\emph{IEEE Transactions on Systems, Man, and
  Cybernetics}} \bibinfo{number}{4} (\bibinfo{year}{1976}),
  \bibinfo{pages}{325--327}.
\newblock


\bibitem[\protect\citeauthoryear{Erfani, Rajasegarar, Karunasekera, and
  Leckie}{Erfani et~al\mbox{.}}{2016}]%
        {erfani2016high}
\bibfield{author}{\bibinfo{person}{Sarah~M Erfani}, \bibinfo{person}{Sutharshan
  Rajasegarar}, \bibinfo{person}{Shanika Karunasekera}, {and}
  \bibinfo{person}{Christopher Leckie}.} \bibinfo{year}{2016}\natexlab{}.
\newblock \showarticletitle{High-dimensional and large-scale anomaly detection
  using a linear one-class SVM with deep learning}.
\newblock \bibinfo{journal}{\emph{Pattern Recognition}}  \bibinfo{volume}{58}
  (\bibinfo{year}{2016}), \bibinfo{pages}{121--134}.
\newblock


\bibitem[\protect\citeauthoryear{Ester, Kriegel, Sander, Xu,
  et~al\mbox{.}}{Ester et~al\mbox{.}}{1996}]%
        {ester1996density}
\bibfield{author}{\bibinfo{person}{Martin Ester}, \bibinfo{person}{Hans-Peter
  Kriegel}, \bibinfo{person}{J{\"o}rg Sander}, \bibinfo{person}{Xiaowei Xu},
  {et~al\mbox{.}}} \bibinfo{year}{1996}\natexlab{}.
\newblock \showarticletitle{A density-based algorithm for discovering clusters
  in large spatial databases with noise}. In
  \bibinfo{booktitle}{\emph{Knowledge Discovery and Data Mining}},
  Vol.~\bibinfo{volume}{96}. \bibinfo{pages}{226--231}.
\newblock


\bibitem[\protect\citeauthoryear{Fan, Miller, Stolfo, Lee, and Chan}{Fan
  et~al\mbox{.}}{2004}]%
        {fan2004using}
\bibfield{author}{\bibinfo{person}{Wei Fan}, \bibinfo{person}{Matthew Miller},
  \bibinfo{person}{Sal Stolfo}, \bibinfo{person}{Wenke Lee}, {and}
  \bibinfo{person}{Phil Chan}.} \bibinfo{year}{2004}\natexlab{}.
\newblock \showarticletitle{Using artificial anomalies to detect unknown and
  known network intrusions}.
\newblock \bibinfo{journal}{\emph{Knowledge and Information Systems}}
  \bibinfo{volume}{6}, \bibinfo{number}{5} (\bibinfo{year}{2004}),
  \bibinfo{pages}{507--527}.
\newblock


\bibitem[\protect\citeauthoryear{Gautam, Balaji, Sudharsan, Tiwari, and
  Ahuja}{Gautam et~al\mbox{.}}{2019}]%
        {gautam2019localized}
\bibfield{author}{\bibinfo{person}{Chandan Gautam}, \bibinfo{person}{Ramesh
  Balaji}, \bibinfo{person}{K Sudharsan}, \bibinfo{person}{Aruna Tiwari}, {and}
  \bibinfo{person}{Kapil Ahuja}.} \bibinfo{year}{2019}\natexlab{}.
\newblock \showarticletitle{Localized multiple kernel learning for anomaly
  detection: one-class classification}.
\newblock \bibinfo{journal}{\emph{Knowledge-Based Systems}}
  \bibinfo{volume}{165} (\bibinfo{year}{2019}), \bibinfo{pages}{241--252}.
\newblock


\bibitem[\protect\citeauthoryear{Gogoi, Borah, and Bhattacharyya}{Gogoi
  et~al\mbox{.}}{2010}]%
        {gogoi2010anomaly}
\bibfield{author}{\bibinfo{person}{Prasanta Gogoi}, \bibinfo{person}{Bhogeswar
  Borah}, {and} \bibinfo{person}{Dhruba~K Bhattacharyya}.}
  \bibinfo{year}{2010}\natexlab{}.
\newblock \showarticletitle{Anomaly detection analysis of intrusion data using
  supervised \& unsupervised approach}.
\newblock \bibinfo{journal}{\emph{Journal of Convergence Information
  Technology}} \bibinfo{volume}{5}, \bibinfo{number}{1} (\bibinfo{year}{2010}),
  \bibinfo{pages}{95--110}.
\newblock


\bibitem[\protect\citeauthoryear{G{\"o}rnitz, Kloft, Rieck, and
  Brefeld}{G{\"o}rnitz et~al\mbox{.}}{2013}]%
        {gornitz2013toward}
\bibfield{author}{\bibinfo{person}{Nico G{\"o}rnitz}, \bibinfo{person}{Marius
  Kloft}, \bibinfo{person}{Konrad Rieck}, {and} \bibinfo{person}{Ulf Brefeld}.}
  \bibinfo{year}{2013}\natexlab{}.
\newblock \showarticletitle{Toward supervised anomaly detection}.
\newblock \bibinfo{journal}{\emph{Journal of Artificial Intelligence Research}}
   \bibinfo{volume}{46} (\bibinfo{year}{2013}), \bibinfo{pages}{235--262}.
\newblock


\bibitem[\protect\citeauthoryear{Gupta, Gao, Aggarwal, and Han}{Gupta
  et~al\mbox{.}}{2014}]%
        {gupta2014outlier}
\bibfield{author}{\bibinfo{person}{Manish Gupta}, \bibinfo{person}{Jing Gao},
  \bibinfo{person}{Charu~C Aggarwal}, {and} \bibinfo{person}{Jiawei Han}.}
  \bibinfo{year}{2014}\natexlab{}.
\newblock \showarticletitle{Outlier detection for temporal data: A survey}.
\newblock \bibinfo{journal}{\emph{IEEE Transactions on Knowledge and Data
  Engineering}} \bibinfo{volume}{26}, \bibinfo{number}{9}
  (\bibinfo{year}{2014}), \bibinfo{pages}{2250--2267}.
\newblock


\bibitem[\protect\citeauthoryear{He, Xu, Huang, and Deng}{He
  et~al\mbox{.}}{2005}]%
        {he2005fp}
\bibfield{author}{\bibinfo{person}{Zengyou He}, \bibinfo{person}{Xiaofei Xu},
  \bibinfo{person}{Zhexue~Joshua Huang}, {and} \bibinfo{person}{Shengchun
  Deng}.} \bibinfo{year}{2005}\natexlab{}.
\newblock \showarticletitle{FP-outlier: Frequent pattern based outlier
  detection}.
\newblock \bibinfo{journal}{\emph{Computer Science and Information Systems}}
  \bibinfo{volume}{2}, \bibinfo{number}{1} (\bibinfo{year}{2005}),
  \bibinfo{pages}{103--118}.
\newblock


\bibitem[\protect\citeauthoryear{Hempstalk, Frank, and Witten}{Hempstalk
  et~al\mbox{.}}{2008}]%
        {hempstalk2008one}
\bibfield{author}{\bibinfo{person}{Kathryn Hempstalk}, \bibinfo{person}{Eibe
  Frank}, {and} \bibinfo{person}{Ian~H Witten}.}
  \bibinfo{year}{2008}\natexlab{}.
\newblock \showarticletitle{One-class classification by combining density and
  class probability estimation}. In \bibinfo{booktitle}{\emph{Proceedings of
  Joint European Conference on Machine Learning and Knowledge Discovery in
  Databases}}.
\newblock


\bibitem[\protect\citeauthoryear{Kannan, Woo, Aggarwal, and Park}{Kannan
  et~al\mbox{.}}{2017}]%
        {Kannan17SDM}
\bibfield{author}{\bibinfo{person}{Ramakrishnan Kannan},
  \bibinfo{person}{Hyenkyun Woo}, \bibinfo{person}{Charu~C Aggarwal}, {and}
  \bibinfo{person}{Haesun Park}.} \bibinfo{year}{2017}\natexlab{}.
\newblock \showarticletitle{Outlier detection for text data}. In
  \bibinfo{booktitle}{\emph{Proceedings of SIAM International Conference on
  Data Mining}}.
\newblock


\bibitem[\protect\citeauthoryear{Keller, M{\"u}ller, and B{\"o}hm}{Keller
  et~al\mbox{.}}{2012}]%
        {Keller2012HiCSHC}
\bibfield{author}{\bibinfo{person}{Fabian Keller}, \bibinfo{person}{Emmanuel
  M{\"u}ller}, {and} \bibinfo{person}{Klemens B{\"o}hm}.}
  \bibinfo{year}{2012}\natexlab{}.
\newblock \showarticletitle{HiCS: High Contrast Subspaces for Density-Based
  Outlier Ranking}. In \bibinfo{booktitle}{\emph{Proceedings of IEEE
  International Conference on Data Engineering}}. \bibinfo{pages}{1037--1048}.
\newblock


\bibitem[\protect\citeauthoryear{Kriegel, Kr\"{o}ger, Schubert, and
  Zimek}{Kriegel et~al\mbox{.}}{2009}]%
        {10.1007/978-3-642-01307-2_86}
\bibfield{author}{\bibinfo{person}{Hans-Peter Kriegel}, \bibinfo{person}{Peer
  Kr\"{o}ger}, \bibinfo{person}{Erich Schubert}, {and} \bibinfo{person}{Arthur
  Zimek}.} \bibinfo{year}{2009}\natexlab{}.
\newblock \showarticletitle{Outlier detection in axis-parallel subspaces of
  high dimensional data}. In \bibinfo{booktitle}{\emph{Proceedings of
  Pacific-Asia Conference on Advances in Knowledge Discovery and Data Mining}}.
\newblock


\bibitem[\protect\citeauthoryear{Kriegel, Schubert, and Zimek}{Kriegel
  et~al\mbox{.}}{2008}]%
        {kriegel2008angle}
\bibfield{author}{\bibinfo{person}{Hans-Peter Kriegel},
  \bibinfo{person}{Matthias Schubert}, {and} \bibinfo{person}{Arthur Zimek}.}
  \bibinfo{year}{2008}\natexlab{}.
\newblock \showarticletitle{Angle-based outlier detection in high-dimensional
  data}. In \bibinfo{booktitle}{\emph{Proceedings of ACM SIGKDD International
  Conference on Knowledge Discovery and Data Mining}}.
  \bibinfo{pages}{444--452}.
\newblock


\bibitem[\protect\citeauthoryear{Kwon, Kim, Kim, Suh, Kim, and Kim}{Kwon
  et~al\mbox{.}}{2017}]%
        {kwon2017survey}
\bibfield{author}{\bibinfo{person}{Donghwoon Kwon}, \bibinfo{person}{Hyunjoo
  Kim}, \bibinfo{person}{Jinoh Kim}, \bibinfo{person}{Sang~C Suh},
  \bibinfo{person}{Ikkyun Kim}, {and} \bibinfo{person}{Kuinam~J Kim}.}
  \bibinfo{year}{2017}\natexlab{}.
\newblock \showarticletitle{A survey of deep learning-based network anomaly
  detection}.
\newblock \bibinfo{journal}{\emph{Cluster Computing}} (\bibinfo{year}{2017}),
  \bibinfo{pages}{1--13}.
\newblock


\bibitem[\protect\citeauthoryear{Lazarevic and Kumar}{Lazarevic and
  Kumar}{2005}]%
        {lazarevic2005feature}
\bibfield{author}{\bibinfo{person}{Aleksandar Lazarevic} {and}
  \bibinfo{person}{Vipin Kumar}.} \bibinfo{year}{2005}\natexlab{}.
\newblock \showarticletitle{Feature bagging for outlier detection}. In
  \bibinfo{booktitle}{\emph{Proceedings of ACM SIGKDD International Conference
  on Knowledge Discovery in Data Mining}}. \bibinfo{pages}{157--166}.
\newblock


\bibitem[\protect\citeauthoryear{Lee, Yeh, and Wang}{Lee et~al\mbox{.}}{2012}]%
        {lee2012anomaly}
\bibfield{author}{\bibinfo{person}{Yuh-Jye Lee}, \bibinfo{person}{Yi-Ren Yeh},
  {and} \bibinfo{person}{Yu-Chiang~Frank Wang}.}
  \bibinfo{year}{2012}\natexlab{}.
\newblock \showarticletitle{Anomaly detection via online oversampling principal
  component analysis}.
\newblock \bibinfo{journal}{\emph{IEEE Transactions on Knowledge and Data
  Engineering}}  \bibinfo{volume}{25} (\bibinfo{year}{2012}),
  \bibinfo{pages}{1460--1470}.
\newblock


\bibitem[\protect\citeauthoryear{Li, Chen, Goh, and Ng}{Li
  et~al\mbox{.}}{2018}]%
        {li2018anomaly}
\bibfield{author}{\bibinfo{person}{Dan Li}, \bibinfo{person}{Dacheng Chen},
  \bibinfo{person}{Jonathan Goh}, {and} \bibinfo{person}{See-Kiong Ng}.}
  \bibinfo{year}{2018}\natexlab{}.
\newblock \showarticletitle{Anomaly detection with generative adversarial
  networks for multivariate time series}.
\newblock \bibinfo{journal}{\emph{arXiv preprint arXiv:1809.04758}}
  (\bibinfo{year}{2018}).
\newblock


\bibitem[\protect\citeauthoryear{Li, Wu, and Du}{Li et~al\mbox{.}}{2017}]%
        {li2017transferred}
\bibfield{author}{\bibinfo{person}{Wei Li}, \bibinfo{person}{Guodong Wu}, {and}
  \bibinfo{person}{Qian Du}.} \bibinfo{year}{2017}\natexlab{}.
\newblock \showarticletitle{Transferred deep learning for anomaly detection in
  hyperspectral imagery}.
\newblock \bibinfo{journal}{\emph{IEEE Geoscience and Remote Sensing Letters}}
  \bibinfo{volume}{14}, \bibinfo{number}{5} (\bibinfo{year}{2017}),
  \bibinfo{pages}{597--601}.
\newblock


\bibitem[\protect\citeauthoryear{Liu, Ting, and Zhou}{Liu
  et~al\mbox{.}}{2008}]%
        {LiuFei2008}
\bibfield{author}{\bibinfo{person}{Fei~Tony Liu}, \bibinfo{person}{Kai~Ming
  Ting}, {and} \bibinfo{person}{Zhi-Hua Zhou}.}
  \bibinfo{year}{2008}\natexlab{}.
\newblock \showarticletitle{Isolation forest}. In
  \bibinfo{booktitle}{\emph{Proceedings of IEEE International Conference on
  Data Mining}}. \bibinfo{pages}{413–422}.
\newblock
\showISBNx{9780769535029}


\bibitem[\protect\citeauthoryear{Liu, Li, Wu, and Fu}{Liu
  et~al\mbox{.}}{2019a}]%
        {liu2019clustering}
\bibfield{author}{\bibinfo{person}{Hongfu Liu}, \bibinfo{person}{Jun Li},
  \bibinfo{person}{Yue Wu}, {and} \bibinfo{person}{Yun Fu}.}
  \bibinfo{year}{2019}\natexlab{a}.
\newblock \showarticletitle{Clustering with outlier removal}.
\newblock \bibinfo{journal}{\emph{IEEE Transactions on Knowledge and Data
  Engineering}} (\bibinfo{year}{2019}).
\newblock


\bibitem[\protect\citeauthoryear{Liu, Zhang, Deng, and Fu}{Liu
  et~al\mbox{.}}{2016}]%
        {liu2016outlier}
\bibfield{author}{\bibinfo{person}{Hongfu Liu}, \bibinfo{person}{Yuchao Zhang},
  \bibinfo{person}{Bo Deng}, {and} \bibinfo{person}{Yun Fu}.}
  \bibinfo{year}{2016}\natexlab{}.
\newblock \showarticletitle{Outlier detection via sampling ensemble}. In
  \bibinfo{booktitle}{\emph{Proceedings of IEEE International Conference on Big
  Data}}.
\newblock


\bibitem[\protect\citeauthoryear{Liu, Li, Zhou, Jiang, Sun, Wang, and He}{Liu
  et~al\mbox{.}}{2019b}]%
        {liu2019generative}
\bibfield{author}{\bibinfo{person}{Yezheng Liu}, \bibinfo{person}{Zhe Li},
  \bibinfo{person}{Chong Zhou}, \bibinfo{person}{Yuanchun Jiang},
  \bibinfo{person}{Jianshan Sun}, \bibinfo{person}{Meng Wang}, {and}
  \bibinfo{person}{Xiangnan He}.} \bibinfo{year}{2019}\natexlab{b}.
\newblock \showarticletitle{Generative adversarial active learning for
  unsupervised outlier detection}.
\newblock \bibinfo{journal}{\emph{IEEE Transactions on Knowledge and Data
  Engineering}} (\bibinfo{year}{2019}).
\newblock


\bibitem[\protect\citeauthoryear{Ma and Perkins}{Ma and Perkins}{2003}]%
        {ma2003time}
\bibfield{author}{\bibinfo{person}{Junshui Ma} {and} \bibinfo{person}{Simon
  Perkins}.} \bibinfo{year}{2003}\natexlab{}.
\newblock \showarticletitle{Time-series novelty detection using one-class
  support vector machines}. In \bibinfo{booktitle}{\emph{Proceedings of
  International Joint Conference on Neural Networks}},
  Vol.~\bibinfo{volume}{3}. \bibinfo{pages}{1741--1745}.
\newblock


\bibitem[\protect\citeauthoryear{Maaten and Hinton}{Maaten and Hinton}{2008}]%
        {maaten2008visualizing}
\bibfield{author}{\bibinfo{person}{Laurens van~der Maaten} {and}
  \bibinfo{person}{Geoffrey Hinton}.} \bibinfo{year}{2008}\natexlab{}.
\newblock \showarticletitle{Visualizing data using t-SNE}.
\newblock \bibinfo{journal}{\emph{Journal of Machine Learning Research}}
  (\bibinfo{year}{2008}), \bibinfo{pages}{2579--2605}.
\newblock


\bibitem[\protect\citeauthoryear{Micenkov{\'a}, McWilliams, and
  Assent}{Micenkov{\'a} et~al\mbox{.}}{2014}]%
        {micenkova2014learning}
\bibfield{author}{\bibinfo{person}{Barbora Micenkov{\'a}},
  \bibinfo{person}{Brian McWilliams}, {and} \bibinfo{person}{Ira Assent}.}
  \bibinfo{year}{2014}\natexlab{}.
\newblock \showarticletitle{Learning outlier ensembles: The best of both
  worlds--supervised and unsupervised}. In
  \bibinfo{booktitle}{\emph{Proceedings of ACM SIGKDD 2014 Workshop on Outlier
  Detection and Description under Data Diversity}}.
\newblock


\bibitem[\protect\citeauthoryear{Pang, Cao, Chen, and Liu}{Pang
  et~al\mbox{.}}{2018}]%
        {pang2018learning}
\bibfield{author}{\bibinfo{person}{Guansong Pang}, \bibinfo{person}{Longbing
  Cao}, \bibinfo{person}{Ling Chen}, {and} \bibinfo{person}{Huan Liu}.}
  \bibinfo{year}{2018}\natexlab{}.
\newblock \showarticletitle{Learning representations of ultrahigh-dimensional
  data for random distance-based outlier detection}. In
  \bibinfo{booktitle}{\emph{Proceedings of the 24th ACM SIGKDD international
  conference on knowledge discovery \& data mining}}.
  \bibinfo{pages}{2041--2050}.
\newblock


\bibitem[\protect\citeauthoryear{Pham and Pagh}{Pham and Pagh}{2012}]%
        {pham2012near}
\bibfield{author}{\bibinfo{person}{Ninh Pham} {and} \bibinfo{person}{Rasmus
  Pagh}.} \bibinfo{year}{2012}\natexlab{}.
\newblock \showarticletitle{A near-linear time approximation algorithm for
  angle-based outlier detection in high-dimensional data}. In
  \bibinfo{booktitle}{\emph{Proceedings of ACM SIGKDD International Conference
  on Knowledge Discovery and Data Mining}}.
\newblock


\bibitem[\protect\citeauthoryear{Rayana}{Rayana}{2016}]%
        {Rayana:2016}
\bibfield{author}{\bibinfo{person}{Shebuti Rayana}.}
  \bibinfo{year}{2016}\natexlab{}.
\newblock \bibinfo{title}{ODDS Library}.
\newblock
\newblock
\urldef\tempurl%
\url{http://odds.cs.stonybrook.edu.}
\showURL{%
\tempurl}


\bibitem[\protect\citeauthoryear{Ren, Zeng, Yang, and Urtasun}{Ren
  et~al\mbox{.}}{2018}]%
        {ren2018learning}
\bibfield{author}{\bibinfo{person}{Mengye Ren}, \bibinfo{person}{Wenyuan Zeng},
  \bibinfo{person}{Bin Yang}, {and} \bibinfo{person}{Raquel Urtasun}.}
  \bibinfo{year}{2018}\natexlab{}.
\newblock \showarticletitle{Learning to reweight examples for robust deep
  learning}.
\newblock \bibinfo{journal}{\emph{arXiv preprint arXiv:1803.09050}}
  (\bibinfo{year}{2018}).
\newblock


\bibitem[\protect\citeauthoryear{Roth}{Roth}{2005}]%
        {roth2005outlier}
\bibfield{author}{\bibinfo{person}{Volker Roth}.}
  \bibinfo{year}{2005}\natexlab{}.
\newblock \showarticletitle{Outlier detection with one-class kernel fisher
  discriminants}. In \bibinfo{booktitle}{\emph{Advances in Neural Information
  Processing Systems}}. \bibinfo{pages}{1169--1176}.
\newblock


\bibitem[\protect\citeauthoryear{Sabokrou, Khalooei, Fathy, and Adeli}{Sabokrou
  et~al\mbox{.}}{2018}]%
        {sabokrou2018adversarially}
\bibfield{author}{\bibinfo{person}{Mohammad Sabokrou},
  \bibinfo{person}{Mohammad Khalooei}, \bibinfo{person}{Mahmood Fathy}, {and}
  \bibinfo{person}{Ehsan Adeli}.} \bibinfo{year}{2018}\natexlab{}.
\newblock \showarticletitle{Adversarially learned one-class classifier for
  novelty detection}. In \bibinfo{booktitle}{\emph{Proceedings of IEEE
  Conference on Computer Vision and Pattern Recognition}}.
  \bibinfo{pages}{3379--3388}.
\newblock


\bibitem[\protect\citeauthoryear{Sathe and Aggarwal}{Sathe and
  Aggarwal}{2016}]%
        {SatheA16}
\bibfield{author}{\bibinfo{person}{Saket Sathe} {and} \bibinfo{person}{Charu~C.
  Aggarwal}.} \bibinfo{year}{2016}\natexlab{}.
\newblock \showarticletitle{{LODES:} Local density meets spectral outlier
  detection}. In \bibinfo{booktitle}{\emph{Proceedings of SIAM International
  Conference on Data Mining}}.
\newblock


\bibitem[\protect\citeauthoryear{Tang, Chen, Fu, and Cheung}{Tang
  et~al\mbox{.}}{2002}]%
        {Tang02PKDD}
\bibfield{author}{\bibinfo{person}{Jian Tang}, \bibinfo{person}{Zhixiang Chen},
  \bibinfo{person}{Ada Wai-Chee Fu}, {and} \bibinfo{person}{David~W Cheung}.}
  \bibinfo{year}{2002}\natexlab{}.
\newblock \showarticletitle{Enhancing effectiveness of outlier detections for
  low density patterns}. In \bibinfo{booktitle}{\emph{Proceedings of
  Pacific-Asia Conference on Knowledge Discovery and Data Mining}}.
\newblock


\bibitem[\protect\citeauthoryear{Wang, Pang, Shen, and Ma}{Wang
  et~al\mbox{.}}{2019}]%
        {wang2019unsupervised}
\bibfield{author}{\bibinfo{person}{Hu Wang}, \bibinfo{person}{Guansong Pang},
  \bibinfo{person}{Chunhua Shen}, {and} \bibinfo{person}{Congbo Ma}.}
  \bibinfo{year}{2019}\natexlab{}.
\newblock \showarticletitle{Unsupervised Representation Learning by Predicting
  Random Distances}.
\newblock \bibinfo{journal}{\emph{International Joint Conference on Artificial
  Intelligence}} (\bibinfo{year}{2019}).
\newblock


\bibitem[\protect\citeauthoryear{Wulsin, Blanco, Mani, and Litt}{Wulsin
  et~al\mbox{.}}{2010}]%
        {wulsin2010semi}
\bibfield{author}{\bibinfo{person}{Drausin Wulsin}, \bibinfo{person}{Justin
  Blanco}, \bibinfo{person}{Ram Mani}, {and} \bibinfo{person}{Brian Litt}.}
  \bibinfo{year}{2010}\natexlab{}.
\newblock \showarticletitle{Semi-supervised anomaly detection for EEG waveforms
  using deep belief nets}. In \bibinfo{booktitle}{\emph{Proceedings of
  International Conference on Machine Learning and Applications}}.
\newblock


\bibitem[\protect\citeauthoryear{Xu, Ricci, Yan, Song, and Sebe}{Xu
  et~al\mbox{.}}{2015}]%
        {xu2015learning}
\bibfield{author}{\bibinfo{person}{Dan Xu}, \bibinfo{person}{Elisa Ricci},
  \bibinfo{person}{Yan Yan}, \bibinfo{person}{Jingkuan Song}, {and}
  \bibinfo{person}{Nicu Sebe}.} \bibinfo{year}{2015}\natexlab{}.
\newblock \showarticletitle{Learning deep representations of appearance and
  motion for anomalous event detection}.
\newblock \bibinfo{journal}{\emph{arXiv preprint arXiv:1510.01553}}
  (\bibinfo{year}{2015}).
\newblock


\bibitem[\protect\citeauthoryear{Yang and Gao}{Yang and Gao}{2013}]%
        {yang2013classification}
\bibfield{author}{\bibinfo{person}{Zeping Yang} {and} \bibinfo{person}{Daqi
  Gao}.} \bibinfo{year}{2013}\natexlab{}.
\newblock \showarticletitle{Classification for imbalanced and overlapping
  classes using outlier detection and sampling techniques}.
\newblock \bibinfo{journal}{\emph{Applied Mathematics \& Information Sciences}}
  \bibinfo{volume}{7}, \bibinfo{number}{1} (\bibinfo{year}{2013}),
  \bibinfo{pages}{375--381}.
\newblock


\bibitem[\protect\citeauthoryear{Yu, Ding, Hu, and Liu}{Yu
  et~al\mbox{.}}{2019}]%
        {yu2019knowledge}
\bibfield{author}{\bibinfo{person}{Weiren Yu}, \bibinfo{person}{Zhengming
  Ding}, \bibinfo{person}{Chunming Hu}, {and} \bibinfo{person}{Hongfu Liu}.}
  \bibinfo{year}{2019}\natexlab{}.
\newblock \showarticletitle{Knowledge Reused Outlier Detection}.
\newblock \bibinfo{journal}{\emph{IEEE Access}} (\bibinfo{year}{2019}),
  \bibinfo{pages}{43763--43772}.
\newblock


\bibitem[\protect\citeauthoryear{Zenati, Foo, Lecouat, Manek, and
  Chandrasekhar}{Zenati et~al\mbox{.}}{2018}]%
        {zenati2018efficient}
\bibfield{author}{\bibinfo{person}{Houssam Zenati},
  \bibinfo{person}{Chuan~Sheng Foo}, \bibinfo{person}{Bruno Lecouat},
  \bibinfo{person}{Gaurav Manek}, {and} \bibinfo{person}{Vijay~Ramaseshan
  Chandrasekhar}.} \bibinfo{year}{2018}\natexlab{}.
\newblock \showarticletitle{Efficient gan-based anomaly detection}.
\newblock \bibinfo{journal}{\emph{arXiv preprint arXiv:1802.06222}}
  (\bibinfo{year}{2018}).
\newblock


\bibitem[\protect\citeauthoryear{Zhang, Hutter, and Jin}{Zhang
  et~al\mbox{.}}{2009}]%
        {Zhang09PKDD}
\bibfield{author}{\bibinfo{person}{Ke Zhang}, \bibinfo{person}{Marcus Hutter},
  {and} \bibinfo{person}{Huidong Jin}.} \bibinfo{year}{2009}\natexlab{}.
\newblock \showarticletitle{A new local distance-based outlier detection
  approach for scattered real-world data}. In
  \bibinfo{booktitle}{\emph{Proceedings of Pacific-Asia Conference on Knowledge
  Discovery and Data Mining}}.
\newblock


\bibitem[\protect\citeauthoryear{Zhao and Fu}{Zhao and Fu}{2015}]%
        {zhao2015dual}
\bibfield{author}{\bibinfo{person}{Handong Zhao} {and} \bibinfo{person}{Yun
  Fu}.} \bibinfo{year}{2015}\natexlab{}.
\newblock \showarticletitle{Dual-regularized multi-view outlier detection}. In
  \bibinfo{booktitle}{\emph{the International Joint Conference on Artificial
  Intelligence}}.
\newblock


\bibitem[\protect\citeauthoryear{Zhao, Hryniewicki, Nasrullah, and Li}{Zhao
  et~al\mbox{.}}{2018}]%
        {DBLP:journals/corr/abs-1812-01528}
\bibfield{author}{\bibinfo{person}{Yue Zhao}, \bibinfo{person}{Maciej~K.
  Hryniewicki}, \bibinfo{person}{Zain Nasrullah}, {and} \bibinfo{person}{Zheng
  Li}.} \bibinfo{year}{2018}\natexlab{}.
\newblock \showarticletitle{{LSCP:} Locally selective combination in parallel
  outlier ensembles}.
\newblock \bibinfo{journal}{\emph{CoRR}} (\bibinfo{year}{2018}).
\newblock
\showeprint[arxiv]{1812.01528.}


\bibitem[\protect\citeauthoryear{Zhao, Nasrullah, and Li}{Zhao
  et~al\mbox{.}}{2019}]%
        {zhao2019pyod}
\bibfield{author}{\bibinfo{person}{Yue Zhao}, \bibinfo{person}{Zain Nasrullah},
  {and} \bibinfo{person}{Zheng Li}.} \bibinfo{year}{2019}\natexlab{}.
\newblock \showarticletitle{PyOD: A Python toolbox for scalable outlier
  detection}.
\newblock \bibinfo{journal}{\emph{Journal of Machine Learning Research}}
  (\bibinfo{year}{2019}).
\newblock
\urldef\tempurl%
\url{http://jmlr.org/papers/v20/19-011.html.}
\showURL{%
\tempurl}


\end{thebibliography}

\end{document}